\documentclass[sigconf]{acmart}
\usepackage{makecell}
\usepackage{amsmath,amsfonts}
\usepackage{algorithmic}
\usepackage{graphicx}
\usepackage{textcomp}
\usepackage{xcolor}

\usepackage{array,multirow}

\usepackage{color, colortbl}
\definecolor{Gray}{gray}{0.8}
\definecolor{lGray}{gray}{0.9}
\usepackage{multicol}

\usepackage{enumitem}

\usepackage[linesnumbered,algoruled,boxed,lined,noend]{algorithm2e}

\SetCommentSty{mycommfont}

\newcommand{\commentRed}[1]{{\color[rgb]{0,0,0}#1}}

\usepackage{comment}
\usepackage{url}

\newtheorem{definition}{Definition}

\copyrightyear{2022}
\acmYear{2022}
\setcopyright{acmlicensed}\acmConference[SIGMOD '22]{Proceedings of the 2022 International Conference on Management of Data}{June 12--17, 2022}{Philadelphia, PA, USA}
\acmBooktitle{Proceedings of the 2022 International Conference on Management of Data (SIGMOD '22), June 12--17, 2022, Philadelphia, PA, USA}
\acmPrice{15.00}
\acmDOI{10.1145/3514221.3526183}
\acmISBN{978-1-4503-9249-5/22/06}

\setcopyright{none}
\begin{document}
\fancyhead{}

\title{dCAM: Dimension-wise Class Activation Map \\for Explaining Multivariate Data Series Classification}

\author{Paul Boniol}
\orcid{0000-0001-8516-0123}

\affiliation{
  \institution{Université Paris Cité\\ \small{boniol.paul@gmail.com}}}

\author{Mohammed Meftah}
\orcid{0000-0001-7636-1572}
\affiliation{
  \institution{EDF R\&D\\ \small{mohammed.meftah@edf.fr}}}

\author{Emmanuel Remy}
\orcid{0000-0003-3423-5919}
\affiliation{
  \institution{EDF R\&D\\ \small{emmanuel.remy@edf.fr}}}

\author{Themis Palpanas}
\orcid{0000-0002-8031-0265}
\affiliation{
 \institution{Université Paris Cité \& IUF\\ \small{themis@mi.parisdescartes.fr}}}

\begin{abstract}
Data series classification is an important and challenging problem in data science. Explaining the classification decisions by finding the discriminant parts of the input that led the algorithm to some decision is a real need in many applications. Convolutional neural networks perform well for the data series classification task; though, the explanations provided by this type of algorithms are poor for the specific case of multivariate data series. Addressing this important limitation is a significant challenge. In this paper, we propose a novel method that solves this problem by highlighting both the temporal and dimensional discriminant information. Our contribution is two-fold: we first describe a convolutional architecture that enables the comparison of dimensions; then, we propose a method that returns dCAM, a Dimension-wise Class Activation Map specifically designed for multivariate time series (and CNN-based models). Experiments with several synthetic and real datasets demonstrate that dCAM is not only more accurate than previous approaches, but the only viable solution for discriminant feature discovery and classification explanation in multivariate time series.
This paper has appeared in SIGMOD'22.
\end{abstract}




\maketitle

\section{Introduction}
\label{sec:intro}

Several applications across many domains 
produce big collections of data series\footnote{A data series, or data sequence, is an ordered sequence of points. If the dimension that imposes the ordering of the sequence is time, then we talk about \emph{time series}. In this paper, we use the terms \emph{sequence}, \emph{data series}, and \emph{time series} interchangeably.}, which need to be processed and analyzed~\cite{DBLP:journals/sigmod/Palpanas15,DBLP:conf/sofsem/Palpanas16,DBLP:journals/sigmod/PalpanasB19,DBLP:journals/dagstuhl-reports/BagnallCPZ19}. 
Typical analysis tasks include pattern matching (or similarity search)~\cite{DBLP:journals/pvldb/EchihabiZPB18,DBLP:journals/pvldb/EchihabiZPB19, c19-isip-Palpanas-isaxfamily, DBLP:journals/tkde/PengFP21, DBLP:conf/sigmod/GogolouTEBP20, messijournal, sing, DBLP:journals/kais/LevchenkoKYAMPS21, seanet, hercules}, classification~\cite{DBLP:journals/datamine/YeK11,DBLP:conf/cikm/SchaferL17, DBLP:journals/datamine/LucasSPOZGPW19, DBLP:journals/datamine/ShifazPPW20, DBLP:journals/datamine/DempsterPW20, DBLP:journals/datamine/TanPW20, DBLP:journals/datamine/SchaferL20, inceptionTime}, clustering~\cite{DBLP:conf/sdm/UlanovaBK15,DBLP:journals/sigmod/PaparrizosG16,DBLP:journals/tods/PaparrizosG17,DBLP:journals/pvldb/PaparrizosF19,DBLP:conf/ijcai/Li0Z19}, anomaly detection~\cite{benchref,DBLP:journals/kais/YankovKR08,DBLP:journals/pvldb/BoniolP20,DBLP:conf/edbt/Gao0B20,normajournal,DBLP:journals/pvldb/BoniolPPF21,distrs2g}, motif discovery~\cite{DBLP:journals/datamine/LinardiZPK20,DBLP:journals/kais/GaoL19,DBLP:journals/tkde/ZhuMK21}, and others~\cite{10.1007/978-3-030-93409-5_40}.

Data series classification is a crucial and challenging problem in data science~\cite{RePEc:wsi:ijitdm:v:05:y:2006:i:04:n:s0219622006002258,esling:hal-01577883}. To solve this task, various data series classification algorithms have been proposed in the past few years~\cite{BagnallBakeOff}, applied on a large number of use cases. 
Standard data series classification methods are based on distances to the instances' nearest neighbors, with k-NN classification (using the Euclidean or Dynamic Time Warping (DTW) distances) being a popular baseline method~\cite{DBLP:journals/corr/abs-1810-07758}. 
Nevertheless, recent works have shown that ensemble methods using more advanced classifiers achieve better performance~\cite{COTE,7837946}. 
Following recent breakthroughs in the computer vision community, new studies successfully propose deep learning methods for data series classification~\cite{DBLP:journals/corr/abs-1809-04356,DBLP:journals/corr/CuiCC16,Zheng2014TimeSC,7870510,guennec2016augmentation,Wang2018MultilevelWD,10.1145/2487575.2487700}, such as Convolutional Neural Network (CNN), Residual Neural Network (ResNet)~\cite{DBLP:journals/corr/WangYO16}, and InceptionTime~\cite{inceptionTime}.

 \begin{figure}[tb]
	\centering
	\includegraphics[scale=0.35]{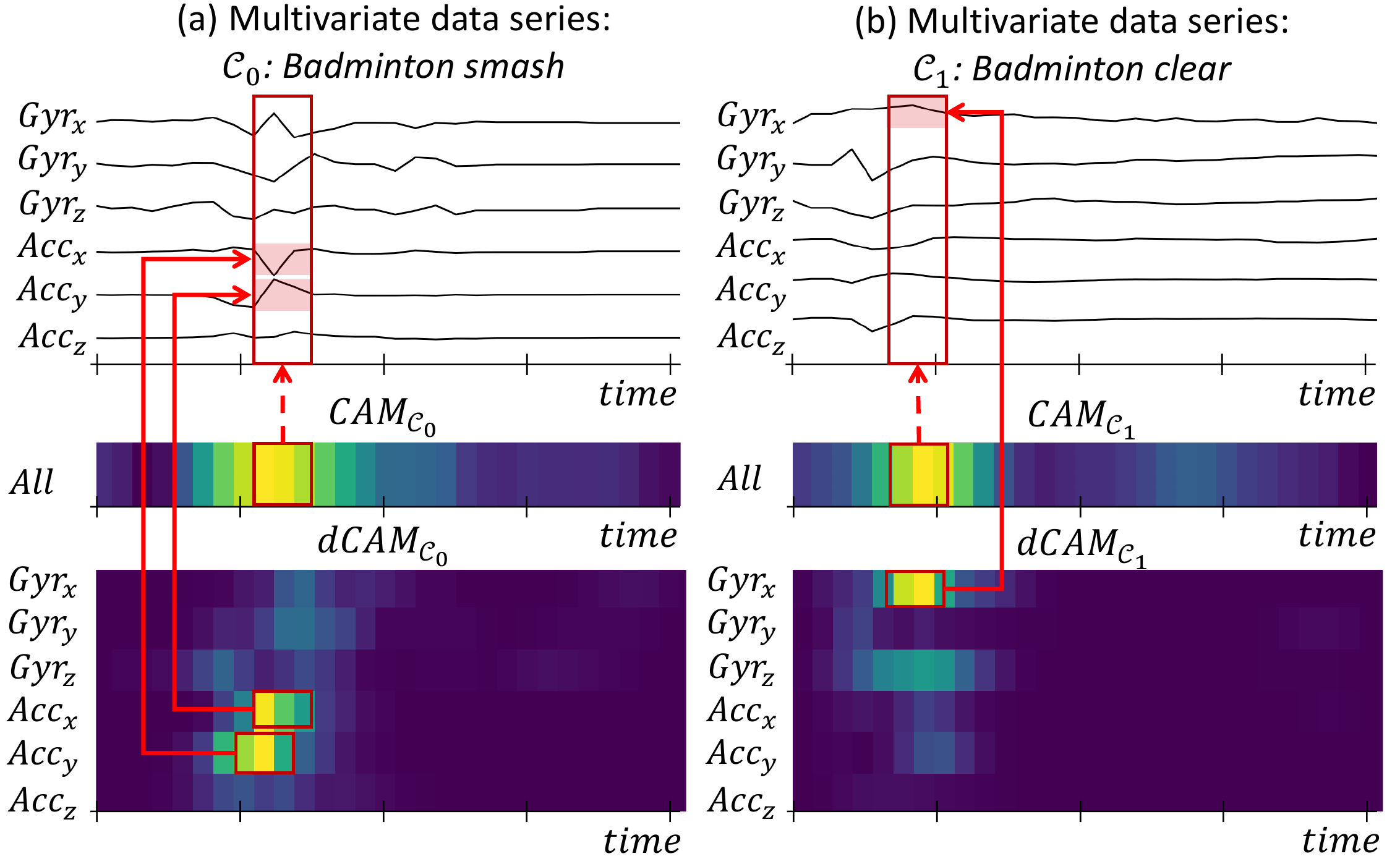}
	\vspace*{-0.4cm}
	\caption{\commentRed{$CAM$ and $dCAM$ computed for two instances (of (a) "badminton smash" and (b) "badminton clear") of RacketSport UCR/UEA dataset}}
	\label{intro_figure}
	\vspace*{-0.5cm}
\end{figure}

\noindent{\bf [Classification Explanation]} While having a trained and accurate classification model, finding explanations of the classification result (i.e., finding the discriminative features that made the model decide which class to attribute to each instance) is a challenging but essential problem, e.g., in manufacturing for anomaly-based predictive maintenance~\cite{Zhang_Song_Chen_Feng_Lumezanu_Cheng_Ni_Zong_Chen_Chawla_2019}, or in medicine for robot-assisted surgeon training~\cite{10.1007/978-3-030-00937-3_25}. 
Such discriminant features can be based on patterns of interest that occur in a subset of dimensions at different timestamps or the same timestamp. 
For 
some CNN-based models, the Class Activation Map (CAM)~\cite{zhou2015cnnlocalization} can be used as an explanation for the classification result. 
CAM has been used for highlighting the parts of an image that contribute the most to a given class prediction and has also been adapted to data series~\cite{DBLP:journals/corr/abs-1809-04356,DBLP:journals/corr/WangYO16}. 

\noindent{\bf [Challenges]} Nevertheless, CAM for data series suffers from one important limitation. 
Since CAM is a univariate time series (of the same length as the input instances) with high values aligned with the subsequences of the input that contribute the most for a given class identification, in the specific case of multivariate data series as input, no information can be retrieved from CAM on the level of contribution of specific dimensions. 
\commentRed{As an example, Figure~\ref{intro_figure} illustrates CAM (top heatmaps) applied on two instances (belonging to two different classes) of the RacketSport UCR/UEA dataset. We observe that CAM explains why the data series correspond to a badminton "smash" or "clear" gesture by highlighting the same temporal window across all dimensions (variables). It is thus not clear what aspect of the gesture distinguishes it from the other.}
Addressing this significant limitation is a sought-after challenge. 

\noindent{\bf [Contributions]} In this paper, we present a novel approach that fills-in the gap by addressing this limitation for the popular CNN-based models. 
We propose a novel data organization and a new CAM technique, dCAM (Dimension-wise Class Activation Map), that is able to highlight both the \emph{temporal and dimensional} information at the same time. 
\commentRed{For instance, in Figure~\ref{intro_figure}, 
dCAM (bottom heatmaps) is pointing to specific subsequences of particular dimensions that explain why the two gestures are different.}
Our method requires only a single training phase, is not constrained by the architecture type, and can efficiently and effectively retrieve discriminant features thanks to a technique that exploits information from different permutations of the input data dimensions. 
Thus, any kind of architecture in which we can apply CAM can benefit from our approach.
Our contributions are as follows. 
\begin{itemize}[leftmargin=*]
\item 
We develop a new method that transforms convolutional-based neural network architectures: whereas previous network architectures can only provide an explanation for all the dimensions together, our transformation represents the only deep learning solution that enables explanation in individual dimensions. 
Our approach can be used with any deep network architecture with a Global Average Pooling layer. 

\item
We demonstrate how we can apply our method to three modern deep learning classification architectures.
We first describe dCNN, inspired by the traditional CNN architecture. 
We then describe how more advanced architectures, such as ResNet and InceptionTime, can be transformed as well. 
We name these transformed architectures dResNet and dInceptionTime. 

\item 
We propose dCAM, a novel method (based on dCNN/ dResNet/dInceptionTime) 
that returns a multivariate CAM, identifying the important parts of the input series for \emph{each} dimension. 

\item
We experimentally demonstrate with several synthetic and real datasets that (among Class Activation Map-based methods) dCAM is not only more accurate in classification than previous approaches, but the only viable solution for discriminant feature discovery and classification explanation in multivariate time series.
Finally, we make our code available online~\cite{ourWebsite}.
\end{itemize}

\section{Background and Related Work}
\label{sec:prelim}

\begin{figure*}
	\centering
	\includegraphics[scale=0.81]{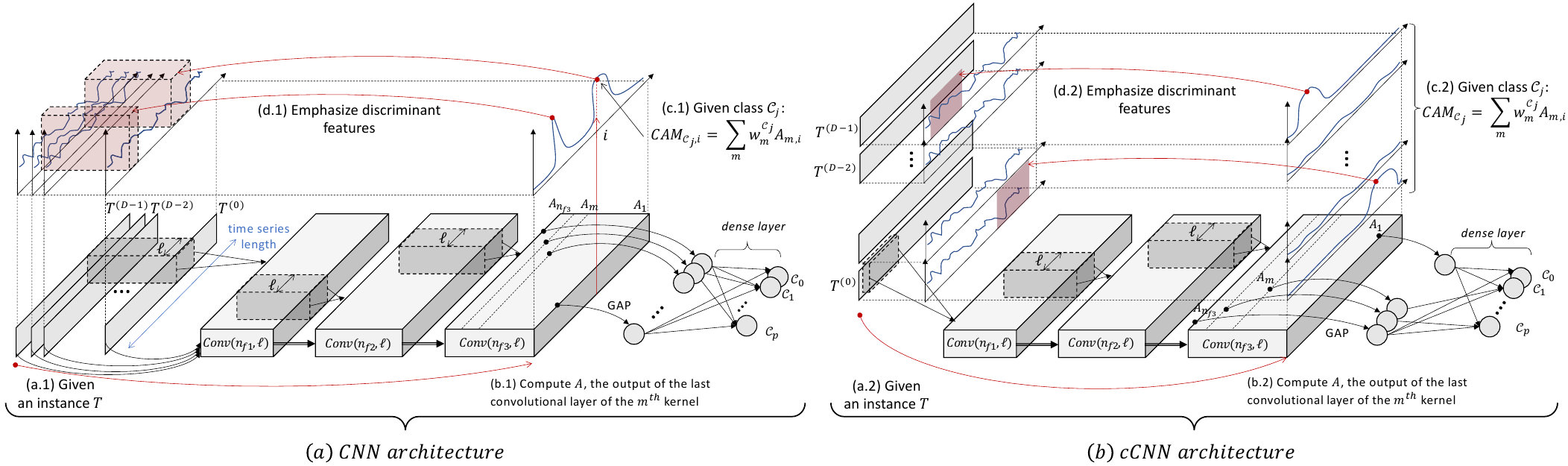}
	\vspace*{-0.3cm}
	\caption{Illustration of Class Activation Map for (a) CNN architecture and (b) cCNN architecture with three convolutional layers ($n_{f1}$, $n_{f2}$, and $n_{f3}$ different kernels respectively of size all equal to $\ell$).}
	\label{CAM_uni}
	\vspace*{-0.3cm}
\end{figure*}

\commentRed{We first present useful notations and definitions, and discuss related work. Table~\ref{SymbolTable} summarizes the symbols we use in this paper.  }

\noindent{\bf [Data Series]}
A multivariate, or $D$-dimensional data series $T \in \mathbb{R}^{(D,n)}$ is a set of $D$ univariate data series of length $n$. We note $T = [T^{(0)}, ... , T^{(D-1)}]$ and for $j \in [0,D-1]$, we note the univariate data series $T^{(j)} = [T^{(j)}_{0}, T^{(j)}_{1}, ... ,T^{(j)}_{n-1}]$.
A subsequence $T^{(j)}_{i,\ell} \in \mathbb{R}^{\ell}$ of the dimension $T^{(j)}$ of the multivariate data series $T$ is a subset of contiguous values from $T^{(j)}$ of length $\ell$ (usually $\ell \ll n$) starting at position $i$; formally, $T^{(j)}_{i,\ell} = [T^{(j)}_{i}, T^{(j)}_{i+1},...,T^{(j)}_{i+\ell-1}]$.

\noindent{\bf [Neural Network Notations]}
We are interested in classifying data series using a neural network architecture model. 
We define the neural network input as $X \in \mathbb{R}^n$ for univariate data series (with $x_i$ the $i^{th}$ value and $X_{i,\ell}$ the sequence of $\ell$ values following the $i^{th}$ value), and $\mathbf{X} \in \mathbb{R}^{(D,n)}$ for multivariate data series (with $x_{j,i}$ the $i^{th}$ value on the $j^{th}$ dimension and $\mathbf{X}_{j,i,\ell}$ the sequence of $\ell$ values following the $i^{th}$ value on the $j^{th}$ dimension).

\noindent\underline{Dense Layer:}
The basic layer of neural network is a fully connected layer (also called $Dense$ $layer$) in which every input neuron is weighted and summed before passing through an activation function. For univariate data series, given an input data series $X \in \mathbb{R}^n$, given a vector of weights $W \in \mathbb{R}^n$ and a vector $B \in \mathbb{R}^n$, we have:
{\small
\begin{align}
\begin{split}
h &= f_{a}\bigg( \sum_{x_i,w_i,b_i \in (X,W,B)}w_i*x_i + b_i \bigg) \\
\end{split}
\label{eq:Dense}
\end{align}
} 
$f_{a}$ is called the activation function and is a non-linear function.
The commonly used activation function $f_{a}$ is the rectified linear unit (ReLU) \cite{Nair:2010:RLU:3104322.3104425} that prevents the saturation of the gradient (other functions that have been proposed are $Tanh$ and Leaky $ReLU$~\cite{DBLP:journals/corr/XuWCL15}). 
For the specific case of multivariate data series, all dimensions are concatenated to give input $X,W \in \mathbb{R}^{D*n}$.
Finally, one can decide to have several output neurons. In this case, each neuron is associated with a different $W$ and $B$, and Equation~\ref{eq:Dense} is executed independently.

\noindent\underline{Convolutional Layer:}
This layer 
has played a significant role for image classification~\cite{Krizhevsky:2012:ICD:2999134.2999257,0483bd9444a348c8b59d54a190839ec9,DBLP:journals/corr/WangYO16}, and recently for data series classification~\cite{DBLP:journals/corr/abs-1809-04356}. 
Formally, for multivariate data series, given an input vector $\mathbf{X} \in \mathbb{R}^{(D,n)}$, and given matrices weights $\mathbf{W},\mathbf{B} \in \mathbb{R}^{(D,\ell)}$, the output $h \in \mathbb{R}^{n}$ of a convolutional layer can be seen as a univariate data series. The tuple $(W,B)$ is also called kernel, with $(D,\ell)$ the size of the kernel. Formally, for $h = [h_0, ... , h_n]$, we have:
{\small
\begin{align}
\begin{split}
h_i &= f_{a} \bigg( \sum_{\substack{X^{(j)},W^{(j)},B^{(j)} \in \\ (\mathbf{X},\mathbf{W},\mathbf{B})}} \sum_{\substack{x_k,w_k,b_k \in \\ (X^{(j)}_{i-\left\lfloor\frac{\ell}{2}\right\rfloor,i+\left\lfloor\frac{\ell}{2}\right\rfloor},W^{(j)},B^{(j)})}} w_k*x_k + b_k \bigg) \\
\end{split}
\end{align}
} 
In practice, we have several kernels of size $(D,\ell)$. 
The result is a multivariate series with dimensions equal to the number of kernels, $n_f$. 
For a given input $\mathbf{X} \in \mathbb{R}^{(D,n)}$, we define $A \in \mathbb{R}^{(n_f,n)}$ to be the output of a convolutional layer $conv(n_f,\ell)$. 
$A_m$ is thus a univariate series corresponding to the output of the $m^{th}$ kernel. 
We denote with $A_m(T)$ the univariate series corresponding to the output of the $m^{th}$ kernel, when $T$ is used as input. 

\noindent\underline{Global Average Pooling Layer:}
Another type of layer 
frequently used 
is pooling. 
Pooling layers compute average/max/min operations, 
aggregating values of previous layers into a smaller number of values for the next layer. 
A specific type of pooling layer is Global Average Pooling (GAP). 
This operation is averaging an entire output of a convolutional layer $A_m(T)$ into one value, 
thus providing invariance to the position of the discriminative features. 

\noindent\underline{Learning Phase:}
The learning phase uses a loss function $\mathcal{L}$ that measures the accuracy of the model and optimizes the various weights. 
For the sake of simplicity, we note $\Omega$ the set containing all weights (e.g., matrices $\mathbf{W}$ and $\mathbf{B}$ defined in the previous sections). Given a set of instances $\mathcal{X}$, we define the average loss as:
$J(\Omega) = \frac{1}{|\mathcal{X}|} \sum_{\mathbf{X} \in \mathcal{X}} \mathcal{L}(\mathbf{X})$.
Then for a given learning rate $\alpha$, the average loss is back-propagated to all weights in the different layers. 
Formally, back-propagation is defined as follows:
$\forall \omega \in \Omega, \omega \leftarrow \omega - \alpha \frac{\partial J}{\partial \omega}$.
In this paper, we use the stochastic gradient descent using the ADAM optimizer~\cite{kingma2017adam} and cross-entropy loss function. 

\subsection{Convolutional-based Neural Network}

We now describe the standard architectures used in the literature.
The first is Convolutional Neural Networks (CNNs)~\cite{DBLP:journals/corr/abs-1809-04356,DBLP:journals/corr/WangYO16}. 
CNN is a concatenation of convolutional layers (joined with $ReLU$ activation functions and batch normalization). 
The last convolutional layer is connected to a Global Average Pooling layer and a dense layer. 
In theory, instances of multiple lengths can be used with the same network. 
A second architecture is the Residual Neural Network (ResNet)~\cite{DBLP:journals/corr/abs-1809-04356,DBLP:journals/corr/WangYO16}. 
This architecture is based on the classical CNN, to which we add residual connections between successive blocks of convolutional layers to avoid that the gradients explode or vanish. 
Other methods have been proposed in the literature~\cite{DBLP:journals/corr/abs-1805-03908,DBLP:journals/corr/abs-1809-04356,DBLP:journals/corr/CuiCC16,inceptionTime}, 
though, CNN and ResNet have been shown to perform the best for multivariate time series classification~\cite{DBLP:journals/corr/abs-1809-04356}. 
InceptionTime~\cite{inceptionTime} has not been evaluated on multivariate data series, but demonstrated state-of-the-art performance on univariate data series. 
\commentRed{Finally, other kinds of architectures than convolutional ones have been proposed in the literature. Attention-based models have been introduced, such as TapNet~\cite{DBLP:conf/aaai/ZhangG0L20}. For the specific case of temporal data, recurrent-based models, such as Recurrent Neural Neworks~\cite{RNNref} (RNN), Long-Short Term Memory~\cite{LSTMref} (LSTM), and Gated Recurrent Unit~\cite{GRUref} (GRU) have received a lot of attention. These three models are relevant to the data series classification task, and we include them in our experimental study.}

\subsection{Class Activation Map (CAM)}

Once the model is trained, we need to find the discriminative features that led the model to decide which class to attribute to each instance. 
\commentRed{Several methods have been proposed to extract meaningful information from CNNs, such as grad-CAM~\cite{8237336} that uses the gradients of the weights to compute the discriminant features, and CAM~\cite{zhou2015cnnlocalization}.}
The latter has been proposed to highlight the parts of an image that contributes the most to a given class identification. The latter has been experimented on data series~\cite{DBLP:journals/corr/abs-1809-04356,DBLP:journals/corr/WangYO16} (univariate and multivariate).
This method explains the classification of a certain deep learning model by emphasizing the subsequences
that contribute the most to a certain classification. 
Note that the CAM method can only be used if (i) a Global Average Pooling layer has been used before the $softmax$ classifier, (ii) the model accuracy is high enough. 
Thus, only the standard architectures CNN and ResNet proposed in the literature can benefit from CAM. 
We now define the CAM method~\cite{DBLP:journals/corr/abs-1809-04356,DBLP:journals/corr/WangYO16}. 
For an input data series $T$, let $A(T)$ be the result of the last convolutional layer $conv(n_f,\ell)$, which is a multivariate data series with $n_f$ dimensions and of length $n$. 
$A_m(T)$ is the univariate time series for the dimension $m \in [1, n_f]$ corresponding to the $m^{th}$ kernel. 
Let $w^{\mathcal{C}_j}_{m}$ be the weight
between the $m^{th}$ kernel and the output neuron of class $\mathcal{C}_j \in \mathcal{C}$. 
Since a Global Average Pooling layer is used, the input to the neuron of class $\mathcal{C}_j$ can be expressed by the following equation:
{\small
\[
z_{\mathcal{C}_j}(T) = \sum_{m} w^{\mathcal{C}_j}_{m} \sum_{A_{m,i}(T) \in A_m(T)} A_{m,i}(T).
\vspace*{-0.1cm}
\]
}
The second sum represents the averaged time series over the whole time dimension. 
Note that weight $w^{\mathcal{C}_j}_{m}$ is independent of index $i$. 
Thus, $z_{\mathcal{C}_j}$ can also be written by the following equation:
{\small
\[
z_{\mathcal{C}_j}(T) = \sum_{A_{m,i}(T) \in A_m(T)}\sum_{m} w^{\mathcal{C}_j}_{m} A_{m,i}(T).
\vspace*{-0.1cm}
\]
}
Finally, $CAM_{\mathcal{C}_j}(T)=[CAM_{\mathcal{C}_j,0}(T), ... , CAM_{\mathcal{C}_j,n-1}(T)]$ that underlines the discriminative features of class $\mathcal{C}_j$ is defined as follows:
{\small
\[
\forall i \in [0,n-1], CAM_{\mathcal{C}_j,i}(T) = \sum_{m} w^{\mathcal{C}_j}_{m} A_{m,i}(T).
\vspace*{-0.1cm}
\]
}
As a consequence, $CAM_{\mathcal{C}_j}(T)$ is a univariate data series where each element at index $i$ indicates the significance of the index $i$ (regardless of the dimensions) for the classification as class $\mathcal{C}_j$. Figure~\ref{CAM_uni}(a) depicts the process of computing CAM and finding the discriminant subsequences in the initial series.


\subsection{CAM Limitations for Multivariate Series}
\label{backprellimitation}
As mentioned earlier, a CAM that highlights the discriminative subsequences of class $\mathcal{C}_j$, $CAM_{\mathcal{C}_j}(T)$, is a univariate data series. 
The information provided by $CAM_{\mathcal{C}_j}(T)$ is sufficient for the case of univariate series classification, but not for 
multivariate series classification. 
Even though the significant temporal index may be correctly highlighted, no information can be retrieved on which dimension is significant or not. 
Solving this serious limitation is a significant challenge in several domains. 
For that purpose, one can propose rearranging the input structure to the network so that the CAM becomes a multivariate data series. 
A new solution would be to decide to use a 2D convolutional neural network with kernel size $(\ell,1)$, such that each kernel slides on each dimension separately. 
Thus, for an input data series $T$, $\mathbf{A}_{m}(T)$ would become a multivariate data series for the variable $m \in [1, n_f]$, and $A^{(d)}_{m}(T) \in \mathbf{A}_m(T)$ would be a univariate time series that would correspond to the dimension $d$ of the initial data series. 
We call this solution \emph{cCNN}, and we use \emph{cCAM} to refer to the corresponding Class Activation Map. 
Figure~\ref{CAM_uni}(b) illustrates cCNN architecture and cCAM.  
Note that if a GAP layer is used, then architectures other than CNN can be used, as well, such as ResNet and InceptionTime. We denote these baselines as \emph{cResNet} and \emph{cInceptionTime}.

Nevertheless, new limitations arise from this solution.
The dimensions are not compared together: 
each kernel of the input layer will take as input only one of the dimensions at a time. 
Thus, features depending on more than one dimension will not be detected. 

Recent studies study the specific case of multivariate data series classification explanation. 
A benchmark study analyzing the saliency/explanation methods for multivariate time series concluded that the explainable methods work better when the multivariate data series is handled as an image~\cite{DBLP:conf/nips/IsmailGBF20}, such as in the cCNN architecture. 
This confirms the need to propose a method specifically designed for multivariate data series. Finally, some recently proposed approaches~\cite{10.1145/3437963.3441815,8970899} address the problems of identifying the discriminant features and discriminant temporal windows independently from one another. 
For instance, MTEX-CNN~\cite{8970899} is an architecture composed of two blocks. 
The 1st block is similar to cCNN. 
The 2nd block consists of merging the results of the 1st block into a 1D convolutional layer, which enables comparing dimensions. 
A variant of CAM~\cite{8237336} is applied to the last convolutional layer of the 1st block in order to find discriminant features for each dimension. 
The discriminant temporal windows are detected with the CAM applied to the last convolutional layer of the 2nd block. 
In practice however, this architecture does not manage to overcome the limitations of cCNN: discriminant features that depend on several dimensions are not correctly identified by MTEX-CNN, which has similar accuracy to cCNN (we elaborate on this in Section~\ref{sec:experiment}). 

In our experimental evaluation, we compare our approach to the MTEX-CNN, cCNN, cResNet and cInceptionTime, and further demonstrate their limitations when addressing the problem at hand.

\section{Problem Formulation}


Given a set $\mathcal{T}$ of multivariate data series $T=\{T^{(0)},T^{(1)},...,T^{(D-1)}\}$ of $D$ dimensions belonging to classes $\mathcal{C}_j \in \mathcal{C}$, and a model $f:\mathcal{T} \rightarrow \mathcal{C}$, we aim to find a function $g(T,f)$ that returns a multivariate series $g(T,f,\mathcal{C}_j)=\{T^{(0)'},T^{(1)'},...,T^{(D-1)'}\}$, in which $T^{(i)'}$ is a series that has high values if the corresponding subsequences in $T_i$ discriminate $T$ of belonging to another class than $\mathcal{C}_j$.

\begin{table}
\centering
\scalebox{0.81}{
\begin{tabular}{|c|c|}
\hline
{\bf Symbol} & {\bf Description} \\
\hline
$T$									& a data series \\
$|T|$									& length of $T$ \\
$T^{(i)}$								& $i^{th}$ dimensions of $T$ \\
$D$ 									& number of dimension\\
$\mathcal{C}$ 							& set of all classes\\
$\mathcal{C}_j$ 						& one class of $\mathcal{C}$\\
$w^{\mathcal{C}_j}_{m}$					& \makecell{weight of connecting the $m^{th}$ convolutional \\ layer and class $\mathcal{C}_j$ neuron}\\
$A_m(T)$ 							& output of the $m^{th}$ convolutional layer for input $T$\\
$z_{\mathcal{C}_j}(T)$ 					& output of $\mathcal{C}_j$ neuron for input $T$\\
$CAM_{\mathcal{C}_j}(T)$ 				& Class Activation Map for class $\mathcal{C}_j$ and input $T$\\
$\Sigma_T$ 							& set of all possible permutations of $T$ dimensions\\
$S^i_T$ 								& \makecell{$T$ with one possible permutation of \\ its dimensions ($S^i_T \in \Sigma_T$)}\\
$k$ 									& number of permutations\\
$n_g$ 								& \makecell{number of permutations that \\ have been correctly classified by the model}\\
\hline
\end{tabular}
} 
\caption{Table of symbols.}
\vspace*{-0.7cm}
\label{SymbolTable}
\end{table}


\begin{figure*}
	\centering
	\includegraphics[scale=1.0]{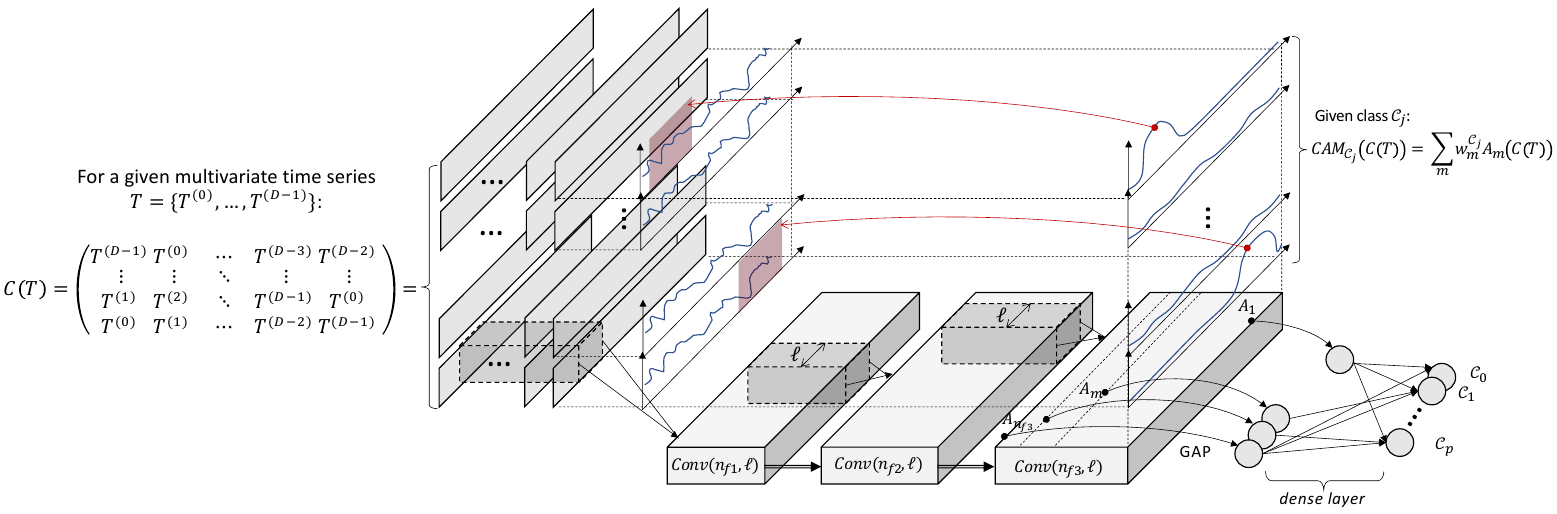}
	\vspace*{-0.3cm}
	\caption{dCNN architecture and application of the CAM.}
	\label{dCNN}
	\vspace*{-0.3cm}
\end{figure*}

\section{Proposed Approach}
\label{sec:solution}

Based on a new architecture that we call dCNN (and variant architectures, e.g., dResNet, dInceptionTime), dCAM aims to provide a multivariate CAM pointing to the discriminant features within each dimension. 
Contrary to the previously described baseline (cCNN, cResNet and cInceptionTime), one kernel on the first convolutional layer will take as input all the dimensions together with different permutations. 
Thus, similarly to the standard CNN architecture, features depending on more than one dimension will be detectable while still having a multivariate CAM. 
Nevertheless, the latter has to be processed such that the significant subsequences are detected. 

We first describe the proposed architecture dCNN that we need in order to provide a dCAM, while still being able to extract multivariate features. 
We then demonstrate that the transformation needed to change CNN to dCNN can also be applied to other, more sophisticated architectures, such as ResNet and InceptionTime, which we denote as dResNet and dInceptionTime. 
We demonstrate that using permutations of the input dimensions makes the classification more robust when important features are localized into small subsequences within some specific dimensions. 

We then present in detail how we compute dCAM (based on a dCNN/dResNet/dInceptionTime architecture). 
Our solution benefits from the permutations injected into the dCNN to identify the most discriminant subsequences used for the classification decision.

\subsection{Dimension-wise Architecture}

As mentioned earlier, the classical CNN architecture mixes all dimensions in the first convolutional layer. 
Thus, the CAM is a univariate data series and does not provide any information on which dimension is the discriminant one for the classification. 
To address this issue, we can use a two-dimensional CNN architecture by re-organizing the input (i.e., the cCNN solution we described earlier). 
In this architecture, one kernel (of size $(1,\ell,1)$) slides on each dimension independently. 
Thus, for a given data series $(T^{(0)}, ... ,T^{(D-1)})$ of length $n$, the convolutional layers returns an array of three dimensions $(n_f,D,n)$, each row $m\in[0,D-1]$ corresponding to the extracted features on dimension $m$. 
Nevertheless, the kernels $(1,\ell,1)$ get as input each dimension independently: 
such an architecture cannot learn features that depend on multiple dimensions.

\subsection{A first Architecture: dCNN}

In order to have the best of both cases, we propose the dCNN architecture, where we transform the input into a cube, in which each row contains a given combination of all dimensions. 
One kernel (of size $(D,\ell,1)$) slides on all dimensions $D$ times. 
This allows the architecture to learn features on multiple dimensions simultaneously. 
Moreover, the resulting CAM is a multivariate data series. 
In this case, one row of the CAM corresponds to a given combination of the dimensions. 
However, we still need to be able to retrieve information for each dimension separately, as well. 
To do that, we make sure that each row contains a different permutation of the dimensions. 
As the weights of the kernels are at fixed positions (for specific dimensions), a permutation of the dimensions will result in a different CAM.
Formally, for a given data series $T$, we note $C(T) \in \mathbb{R}^{(D,D,n)}$ the input data structure of dCNN: 
{\small
\[
C(T) = 
\begin{pmatrix}
T^{(D-1)} & T^{(0)} & ... & T^{(D-3)} & T^{(D-2)}\\
: & : & ... & : & :\\
T^{(1)} & T^{(2)} & ... & T^{(D-1)} & T^{(0)}\\
T^{(0)} & T^{(1)} & ... & T^{(D-2)} & T^{(D-1)}\\
\end{pmatrix}
\]
} 
Note that each row and column of $C(T)$ contains all dimensions. 
Thus, a given dimension $T^{(i)}$ is never at the same position in $C(T)$ rows. 
The latter is a crucial property for the computation of dCAM.
In practice, we guarantee the latter property by shifting by one position the order of the dimensions. Thus $T^{(0)}$ in the first row is aligned with $T^{(1)}$ in the second row. A different order of $T$ dimensions will thus generate a different matrix $C(T)$.

Figure~\ref{dCNN} depicts the dCNN architecture. 
The input $C(T)$ is forwarded into a classical two-dimensional CNN. 
The rest of the architecture is independent of the input data structure. 
The latter means that any other two-dimensional architecture (containing a Global Average Pooling) can be used (such as ResNet), by only adapting the input data structure. 
Similarly, the training procedure can be freely chosen by the user. For the rest of the paper, we will use the cross-entropy loss function and the ADAM optimizer.

Observe that multiple permutations of the original multivariate series will be processed by several convolutional layers, enabling the kernel to examine multiple different combinations of dimensions and subsequences. 
Note that the kernels of the dCNN will be sparse, which has a significant impact on overfitting.

\subsection{The dResNet/dInceptionTime Architectures}

As mentioned earlier, any architecture using a GAP layer after the last convolutional layer can benefit from dCAM. 
Thus, different (and more sophisticated) architectures can be used with our approach. 
To that effect, we propose two new architectures dResNet and dInceptionTime, based on the state-of-the-art architectures ResNet~\cite{DBLP:journals/corr/WangYO16} and InceptionTime~\cite{inceptionTime}. 
The transformations that lead to dResNet and dInceptionTime are very similar to that from CNN to dCNN, using $C(T)$ as input to the transformed networks.
The convolutional layers are transformed from 1D (as originally proposed~\cite{DBLP:journals/corr/WangYO16,inceptionTime}) to 2D. 
Similarly to dCNN, the kernel sizes are $(D,\ell,1)$ and convolute over each row of $C(T)$ independently.

We demonstrate in the experimental section that these architectures do not affect the usage of our proposed approach dCAM, and we evaluate the choice of architecture on both classification and discriminant features identification. 


\subsection{Dimension-wise Class Activation Map}

At this point, we have our network trained to classify instances among classes $\mathcal{C}_0, \mathcal{C}_1, ... ,\mathcal{C}_p$. 
We now explain how to compute dCAM that will identify discriminant features within dimensions.
We assume that the network has to be accurate enough in order to provide a meaningful dCAM. We evaluate in the experimental section the relation between the classification accuracy of the network and the discriminant features identification accuracy of dCAM.

At first glance, we can compute the regular CAM $CAM_{\mathcal{C}_j}(C(T)) = \sum_{m} w_m^{\mathcal{C}_j} A_m(C(T))$. 
However, a high value on the $i^{th}$ row at position $t$ on $CAM_{\mathcal{C}_j}(C(T))$ does not mean that the subsequence at position $t$ on the $i^{th}$ dimension is important for the classification. 
It instead means that the combination of dimensions at the $i^{th}$ row of $C(T)$ is important.

\subsubsection{Random Permutation Computations}

Given those different combinations of dimensions (i.e., one row of $C(T)$) produce different outputs (i.e., the same row in $CAM_{\mathcal{C}_j}(C(T))$), the positions of the dimensions within the $C(T)$ rows have an impact on the CAM. 
Consequently, for a given combination of dimensions, we can assume that at least one dimension at a given position is responsible for the high value in the CAM row.
For the rest of this paper, we use $\Sigma_T$ as the set of all possible permutations of $T$ dimensions, and $S^i_T \in \Sigma_T$ for a single permutation of $T$.
E.g., for a given data series $T=\{T^{(0)},T^{(1)},T^{(2)}\}$, one possible permutation is $S^i_T=\{T^{(1)},T^{(0)},T^{(2)}\}$. 

\begin{figure}
	\centering
	\includegraphics[scale=0.66]{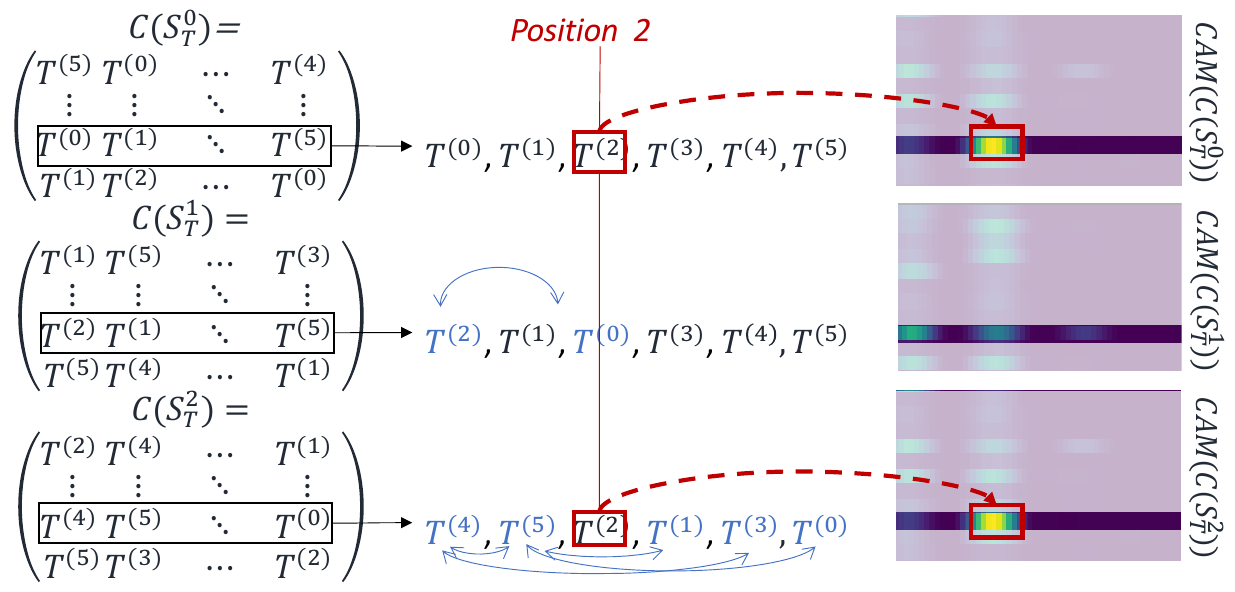}
	\vspace*{-0.4cm}
	\caption{Example of Class Activation Map results for different permutations.}
	\label{Permutation_example}
	\vspace*{-0.3cm}
\end{figure}

Figure~\ref{Permutation_example} depicts an example of CAMs for different permutations. 
In this Figure, for three given permutations of $T$ (i.e., $S^0_T$, $S^1_T$ and $S^2_T$), we notice that when $T^{(2)}$ is in position two of the second row of $C(S^i_T)$, the Class Activation Map $CAM(C(S^i_T))$ is greater than when $T^{(2)}$ is not in position two. 
We infer that the second dimension of $T$ in position two is responsible for the high value.
Thus, we may examine different dimension combinations by keeping track of which dimension at which position is activating the CAM the most. 
We now describe the steps necessary to retrieve this information.


\begin{definition}
For a given data series $T=\{T^{(0)},T^{(1)},...,T^{(D-1)}\}$ of length $n$ and its input data structure $C(T)$, we define function $idx$, such that $idx(T^{(i)},p_j)$ returns the row indices in $C(T)$ that contain the dimension $T^{(i)}$ at position $p_j$.
\end{definition}

We can now define the following transformation $\mathcal{M}$.

\begin{definition}
	\label{def:m}
For a given data series $T=\{T^{(0)},T^{(1)},...,T^{(D-1)}\}$ of length $n$, a given class $\mathcal{C}_j$ and Class Activation Map, we define $\mathcal{M}(CAM_{\mathcal{C}_j}(C(T))) \in \mathbb{R}^{(D,D,n)}$ (with $CAM_{\mathcal{C}_j}(C(T)) \in \mathbb{R}^{(D,n)}$ and $CAM_{\mathcal{C}_j}(C(T))_i$ its $i^{th}$ row) as follows:
{\scriptsize
\begin{multline}
\hspace*{-0.4cm}
\mathcal{M}(CAM_{\mathcal{C}_j}(C(T))) = \\
\begin{pmatrix}
CAM_{\mathcal{C}_j}(C(T))_{idx(T^{(0)},0)} & ... & CAM_{\mathcal{C}_j}(C(T))_{idx(T^{(0)},D-1)}\\
CAM_{\mathcal{C}_j}(C(T))_{idx(T^{(1)},0)} & ... & CAM_{\mathcal{C}_j}(C(T))_{idx(T^{(1)},D-1)}\\
: & ... & :\\
CAM_{\mathcal{C}_j}(C(T))_{idx(T^{(D-1)},0)} & ... & CAM_{\mathcal{C}_j}(C(T))_{idx(T^{(D-1)},D-1)}\\
\end{pmatrix}
\end{multline}
} 
\end{definition}

\begin{figure}
	\centering
	\includegraphics[scale=0.79]{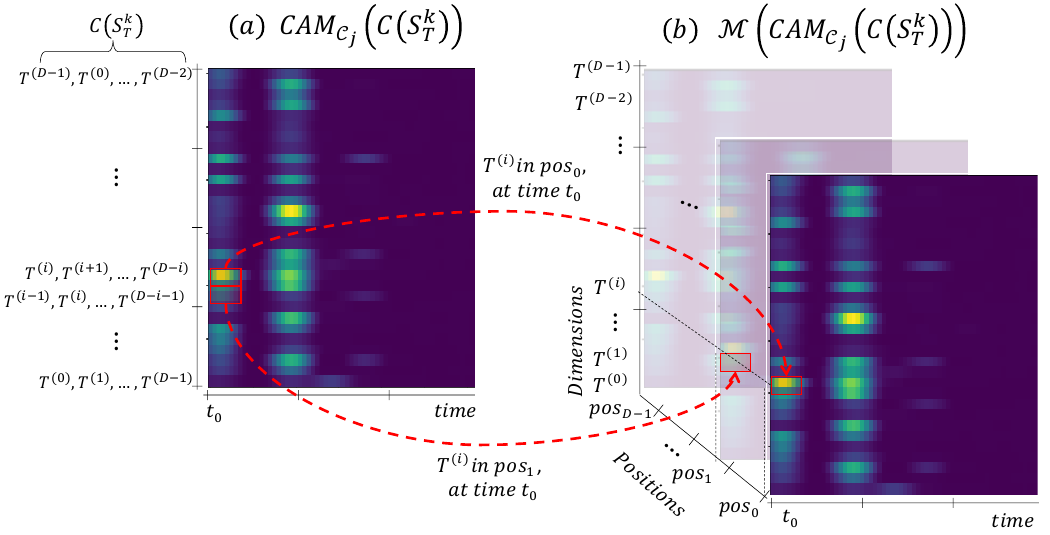}
	\vspace*{-0.7cm}
	\caption{Transformation $\mathcal{M}$ for a given data series $T$.}
	\label{Mfunction}
	\vspace*{-0.3cm}
\end{figure}

Figure~\ref{Mfunction} depicts the $\mathcal{M}$ transformation. 
As explained in Definition~\ref{def:m}, the $\mathcal{M}$ transformation enriches the Class Activation Map by adding the dimension position information. 
Note that if we change the dimension order of $T$, their $\mathcal{M}(CAM_{\mathcal{C}_j}(C(T)))$ changes as well. 
Indeed, for a given dimension $T^{(i)}$ and position $p_j$, $idx(T^{(i)},p_j)$ will not have the same value for two different dimension orders of $T$. 
Thus, computing $\mathcal{M}(CAM_{\mathcal{C}_j}(C(T)))$ for different dimension orders of $T$ will provide distinct information regarding the importance of a given position (subsequence) in a given dimension. 
We expect that subsequences (of a specific dimension) that discriminate one class from another will also be associated (most of the time) with a high value in the Class Activation Map. 

\begin{figure*}
	\centering
	\includegraphics[scale=1.15]{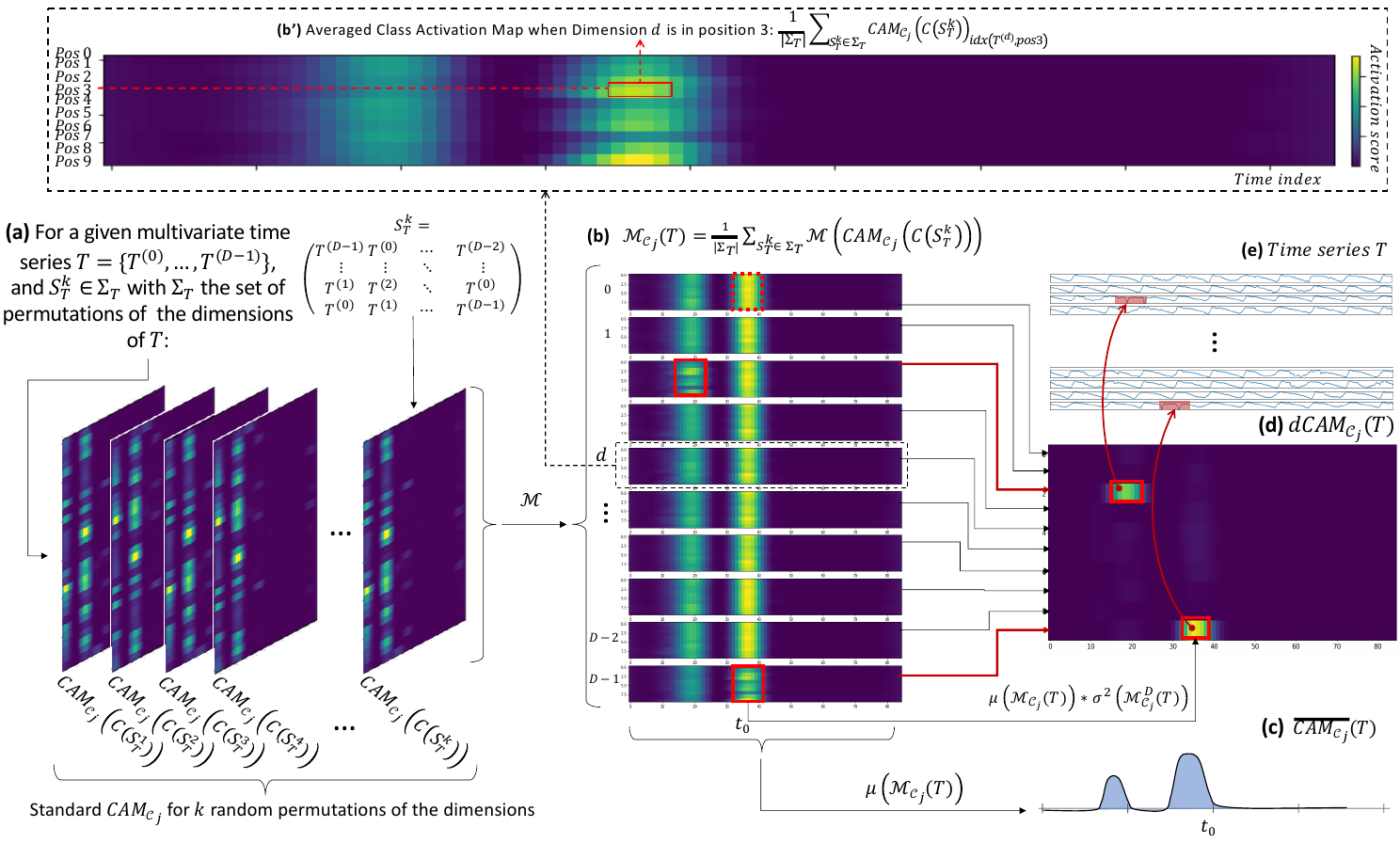}
	\vspace*{-0.7cm}
	\caption{dCAM computation framework.}
	\label{dCAM}
	\vspace*{-0.3cm}
\end{figure*}

\subsubsection{Merging Permutations}


We compute $\mathcal{M}(CAM_{\mathcal{C}_j}(C(S_T)))$, for different $S_T\in\Sigma_T$.
\commentRed{Note that the total number of permutations for high-dimensional data series is enormous:  $|\Sigma_T| =  D!$}. 
In practice, we only compute $\mathcal{M}$ for a randomly selected subset of $\Sigma_T$. 
We thus merge $k=|\Sigma_T|$ permutations $S^k_T$, by computing the averaged matrix $\bar{\mathcal{M}}_{\mathcal{C}_j}(T)$ of all the $\mathcal{M}$ transformations of the permutations:  
{\small
\[
\bar{\mathcal{M}}_{\mathcal{C}_j}(T) = \frac{1}{|\Sigma_T|} \sum_{S^k_T \in \Sigma_T} \mathcal{M}(CAM_{\mathcal{C}_j}(C(S^k_T))) 
\]
\vspace*{-0.2cm}
} 

Figure~\ref{dCAM} illustrates the process of computing $\bar{\mathcal{M}}_{\mathcal{C}_j}(T)$ from the set of permutations of $T$, $\Sigma_T$. $\bar{\mathcal{M}}_{\mathcal{C}_j}(T)$ can be seen as a summarization of the importance of each dimension at each position in $C(T)$, for all the computed permutations. 
Figure~\ref{dCAM}(b') (at the top of the figure) depicts $\bar{\mathcal{M}}_{\mathcal{C}_j}(T)_d$, which corresponds to the $d^{th}$ row (i.e., the dotted box in Figure~\ref{dCAM}(b)) of $\bar{\mathcal{M}}_{\mathcal{C}_j}(T)$. 
Each row of $\bar{\mathcal{M}}_{\mathcal{C}_j}(T)_d$ corresponds to the average activation of dimension $d$ (for each timestamp) when dimension $d$ is in a given position in $C(T)$.

Note that all permutations of $T$ are forwarded into the dCNN network without training it again. Thus, even though the permutations of $T$ generate radically different inputs to the network, the network can still classify most of the instances correctly. 
For $k$ permutations, we use $n_g$ to denote the number of permutations that the model has correctly classified. 
We provide an analysis (see Section~\ref{sec:experiment}) of $n_g/k$ w.r.t the classification accuracy of the model and the impact that $n_g/k$ has on the discriminant features identification accuracy.


\subsubsection{dCAM Extraction}

We can now use the previously computed $\bar{\mathcal{M}}_{\mathcal{C}_j}$ to extract explanatory information on which subsequences are considered important by the network. 
First, we note that each row of $C(T)$ corresponds to the input format of the standard CNN architecture. 
Thus, we expect that the result of a row of $\bar{\mathcal{M}}_{\mathcal{C}_j}$ (one of the ten lines in Figure~\ref{dCAM}(b)) is similar to the standard CAM. 
Hence, we can assume that $\mu(\mathcal{\bar{M}}_{\mathcal{C}_j}(T)) = \sum_{d \in [0,D-1]} \sum_{p \in [0,D-1]} \bar{\mathcal{M}}_{\mathcal{C}_j}^{d,p}(T)/(2*D)$ is equivalent to standard Class Activation Map $CAM_{\mathcal{C}_j}(T)$ (this approximation is depicted in Figure~\ref{dCAM}(d)).
Moreover, in addition to the temporal information, we can extract temporal information per dimension. 
We know that for a given position $p$ and a given dimension $d$, $\bar{\mathcal{M}}_{\mathcal{C}_j}^{d,p}(T)$ represents the averaged activation for a given set of permutations. 
If the activation $\bar{\mathcal{M}}_{\mathcal{C}_j}^{d,p}(T)$ for a given dimension is constant (regardless of its value, or the position $p$), then the position of dimension $d$ is not important, and no subsequence in that dimension $d$ is discriminant. 
On the other hand, a high or low value at a specific position $p$ means that the subsequence at this specific position is discriminant. 
While it is intuitive to interpret a high value, interpreting a low value is counterintuitive. 
Usually, a subsequence at position $p$ with a low value should be regarded as non-discriminant. 
Nevertheless, if the activation is low for $p$ and high for other positions, then the subsequence at position $p$ is the consequence of the low value and is thus discriminant.
We experimentally observe this situation, where a non-discriminant dimension has a constant activation per position (e.g., see dotted red rectangle in Figure~\ref{dCAM}(b): this pattern corresponds to a non-discriminant subsequence of the dataset). 
On the contrary, for discriminant dimensions, we observe a strong variance for the activation per position: either high values or low values (e.g., see solid red rectangles in Figure~\ref{dCAM}(b): these patterns correspond to the (injected) discriminant subsequences highlighted in red in Figure~\ref{dCAM}(e)). 
We thus can extract the significant subsequences per dimension by computing the variance of all positions of a given dimension. 
We can filter out the irrelevant temporal windows using the averaged $\mu(\mathcal{\bar{M}}_{\mathcal{C}_j}(T))$ for all dimensions, and use the variance to identify the important dimensions in the relevant temporal windows. 
Formally, 
we define $dCAM_{\mathcal{C}_j}(T)$ as follows.

\begin{definition}
For a 
data series $T$ and 
class $\mathcal{C}_i$, $dCAM_{\mathcal{C}_j}(T)$ is: 
\begin{footnotesize}
\begin{multline}
dCAM_{\mathcal{C}_j}(T) = \\
\begin{pmatrix}
\sigma^2(\bar{\mathcal{M}}_{\mathcal{C}_j}^{0}(T)_{t_0})*\mu(\bar{\mathcal{M}}_{\mathcal{C}_j}(T)_{t_0}) & ... & \sigma^2(\bar{\mathcal{M}}_{\mathcal{C}_j}^{0}(T)_{t_n})*\mu(\bar{\mathcal{M}}_{\mathcal{C}_j}(T)_{t_n})\\
: & ... & :\\
\sigma^2(\bar{\mathcal{M}}_{\mathcal{C}_j}^{D-2}(T)_{t_{0}})*\mu(\bar{\mathcal{M}}_{\mathcal{C}_j}(T)_{t_0}) & ... & \sigma^2(\bar{\mathcal{M}}_{\mathcal{C}_j}^{D-2}(T)_{t_{n}})*\mu(\bar{\mathcal{M}}_{\mathcal{C}_j}(T)_{t_n})\\
\sigma^2(\bar{\mathcal{M}}_{\mathcal{C}_j}^{D-1}(T)_{t_0})*\mu(\bar{\mathcal{M}}_{\mathcal{C}_j}(T)_{t_0}) & ... & \sigma^2(\bar{\mathcal{M}}_{\mathcal{C}_j}^{D-1}(T)_{t_n})*\mu(\bar{\mathcal{M}}_{\mathcal{C}_j}(T)_{t_n})\\
\end{pmatrix}
\end{multline}
\vspace*{-0.3cm}
\end{footnotesize}
\end{definition}

\commentRed{
\subsection{Time Complexity Analysis}
 
 
\noindent {\bf[Training step]}: CNN/ResNet/InceptionTime require $O(\ell*|T|*D)$ computations per kernel, while  dCNN/dResNet/dInceptionTime require $O(\ell*|T|*D^2)$ computations per kernel. 
Thus, the training time per epoch is higher for dCNN than CNN. However, given that the size of the input of dCNN is larger (containing $D$ permutations of a single series) than CNN, the number of epochs to reach convergence is lower for dCNN when compared to CNN.  Intuitively, dCNN trains on more data during a single epoch. 
This leads to similar overall training times (see Section~\ref{sc:executiontime}). 

\noindent {\bf[dCAM step]}: 
The CAM computation complexity is $O(|T|*D*n_f)$, where $n_f$ is the number of filters in the last convolutional layer.
Let $N_f = [n_{f_1}, ..., n_{f_n}]$ be the number of filters of the $n$ convolutional layers. 
Then, a forward pass has time complexity $O(\ell*|T|*D^2*\sum_{n_{f_i} \in N_f} n_{f_i})$. 
In dCAM, we evaluate $k$ different permutations. 
Thus, the overall dCAM complexity is $O(k*\ell*|T|*D^2*\sum_{n_{f_i} \in N_f} n_{f_i})$. 
Observe that since the $k$ permutations can be computed in parallel, the most important parameter for the execution time is $D$.
}

\commentRed{
\subsection{Further Observations}
\label{sec:discussion}
%
We note that since in real use cases, labels are not available, 
the number of correctly classified permutations (called $n_g$) could be used as a proxy to assess the quality of the explanation (see Section~\ref{sc:parameteranalysis}). 
%

Moreover, when analyzing sets of series, we can use dCAM on each one independently, and then aggregate the dCAM results to identify global discriminant features (see Section~\ref{sec:surgery}). 
}

%

 
\section{Experimental Evaluation}
\label{sec:experiment}

\subsection{Experimental Setup}
\label{sc:expsetup}
\commentRed{We implemented our algorithms in Python 3.5 using the PyTorch library~\cite{NEURIPS2019_bdbca288}.} 
The evaluation was conducted on a server with Intel Core i7-8750H CPU 2.20GHz x 12, with 31.3GB RAM, and Quadro P1000/PCle/SSE2 GPU with 4.2GB RAM, and on Jean Zay cluster with Nvidia Tesla V100 SXM2 GPU with 32 GB RAM.

Our code and datasets are available online~\cite{ourWebsite}.

\subsubsection{{\bf Datasets}}

We conduct our experimental evaluation using real datasets from the UCR/UEA archive~\cite{DBLP:journals/corr/abs-1810-07758} to evaluate the classification performance of the competing methods.
%
The real datasets are injected with known discriminant patterns and a real use case from the medical domain to evaluate the discriminant features identification. 
We use the StarLightCurves (classes 2 and 3 only), ShapesAll (classes 1 and 2 only), and Fish (class 1 and 2 only) datasets from the UCR archive~\cite{DBLP:journals/corr/abs-1810-07758}, in which we inject subsequences that will generate discriminant features. 
We build two types of datasets to study the ability of the algorithms to identify the discriminant patterns guiding the classification decision, (1) when these patterns occur in a subset of the dimensions at different timestamps, and (2) when these patterns occur in a subset of the dimensions at the same timestamp. 

 \begin{figure*}
	\centering
	\includegraphics[scale=1.14]{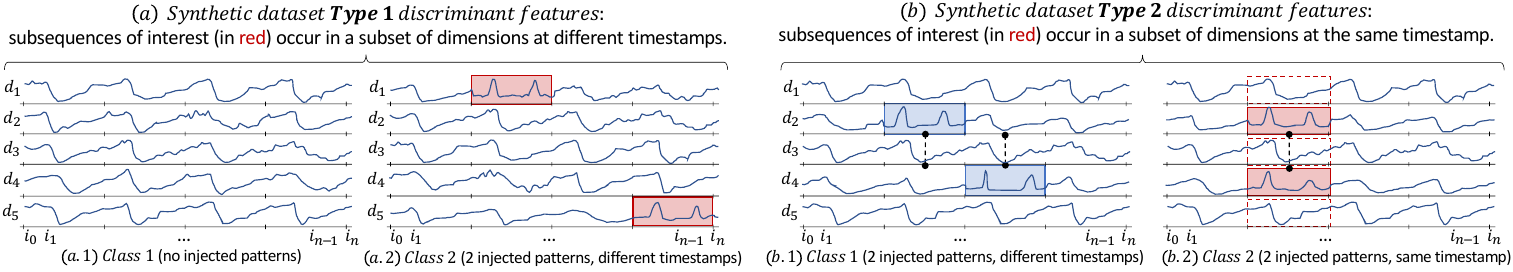}
	\vspace*{-0.4cm}
	\caption{Synthetic datasets: (a) $Type$ $1$, in which the discriminant subsequence is two injected patterns from class 2 StarLightCurves dataset in random dimensions at random positions, (b) $Type$ $2$, in which the discriminant factor is the fact that the two injected patterns are injected at the same position.}
	\label{Synthetic}
	\vspace*{-0.2cm}
\end{figure*}

\noindent (1) For the $Type$ $1$ datasets, we build each dimension of Class 1 by concatenating random instances from one class of one of our two UCR seed datasets. 
We build Class 2 by injecting in the series of the other class of our two UCR datasets a pattern in $2$ random dimensions at a random position in the series.

\noindent (2) For the $Type$ $2$ datasets, we build each dimension of Class 1 by concatenating random instances from one of the classes of our two UCR datasets and injecting patterns from the other class in $x$ random dimensions and at different positions. 
We build Class 2 by injecting patterns at the same positions of $2$ random dimensions. 

Examples of $Type$ $1$ and $Type$ $2$ 5-dimensional datasets based on StarLightCurves are depicted in Figures~\ref{Synthetic}(a), and~\ref{Synthetic}(b), respectively; 
we use 1000 such datasets.
In addition, we consider a use case from medicine related to robot-assisted surgeon training ( Section~\ref{sec:surgery}).

\subsubsection{{\bf Evaluation Measures}}
We first evaluate the classification accuracy, $C$-$acc$. 
This measure corresponds to the ratio of correctly classified instances among all instances in the test dataset. 

We then evaluate the discriminant features accuracy, $Dr$-$acc$, for Class 1 (see Figure~\ref{Synthetic}).
We define $Dr$-$acc$ as the PR-AUC for CAM/cCAM/dCAM obtained from the models and the ground-truth. 
The ground-truth is a series that has $1$ at the positions of discriminant features (see Figure~\ref{Synthetic}(a.2): ground-truth contains $1$ at the positions of the injected patterns, marked with the red rectangles, and $0$ otherwise). 
We motivate the choice of PR-AUC (instead of ROC-AUC) because we are more interested in measuring the accuracy of identifying the injected patterns (representing at max 0.02 percent of the dataset) than measuring the accuracy of not detecting the non-injected patterns. 
In this very unbalanced case, PR-AUC is more appropriate than ROC AUC~\cite{10.1145/1143844.1143874}.

Note that even though we annotate each point of the injected subsequences as discriminant, only some subparts of these sequences may be discriminant, thus, leading to $Dr$-$acc$ less than $1$. 
Finally, for CNN/ResNet/InceptionTime, we compute the $Dr$-$acc$ scores by assuming that their (univariate) CAM values are the same for all dimensions.
We mark their $Dr$-$acc$ scores with a star in Table~\ref{result_Multi_Normal}.

\subsection{{\bf Baselines and Training Setup}}
We compare our model, dCNN/dResNet/dInceptionTime, to the classical CNN/ResNet/InceptionTime model~\cite{10.1007/978-3-030-00937-3_25,DBLP:journals/corr/abs-1809-04356,DBLP:journals/corr/WangYO16,inceptionTime}, and the cCNN/cResNet/cInceptionTime baseline we introduced in Section~\ref{sec:prelim}.
We are using the same architecture setup for all models. 
We then use CAM for CNN, ResNet, InceptionTime, cCAM for cCNN, cResNet and cInceptionTime and dCAM for dCNN, dResNet and dInceptionTime to identify discriminant features.
For CNN, cCNN and dCNN, we are using 5 convolutional layers with $(64,128,256,256,256)$ filters respectively. We are using a kernel size of 3 and a padding of 2.
For ResNet, cResNet, and dResNet, we are using three blocks with three convolutional layers of 64 filters (for the first two blocks) and 128 layers (for the last block). 
We are using kernel sizes equal to 8, 5, and 3 for each block for the three layers of the block.
For InceptionTime, cInceptionTime and dInceptionTime, we are using the same architecture as originally defined~\cite{inceptionTime}.
 
We also include MTEX-CNN~\cite{8970899}(MTEX) as a baseline, representative of other kinds of architectures that can provide a multivariate CAM. 
The explanation is computed separately for discriminant features and timestamps using grad-CAM~\cite{8237336} (MTEX-grad). The latter is a variant of the usual CAM \commentRed{using the gradients of the weights instead of the GAP layer to compute the activation.}
 
\commentRed{We finally include three recurrent neural networks: the usual Recurrent Neural Network~\cite{RNNref} (RNN), Long-Short Term Memory~\cite{LSTMref} (LSTM), and Gated Recurrent Unit~\cite{GRUref} (GRU) to our benchmark. As following previous evaluation work conducted in the UCR/UEA archive~\cite{RNNEXPref}, we use for all networks one recurrent hidden layer (RNN, LSTM, and GRU respectively) of 128 neurons. We then add one dense layer connecting the 128 neurons to the classes neurons.}
 
We split our dataset into training and validation sets with 80 and 20 percent of the total dataset, respectively (equally balanced between the two classes). The training dataset is used to train the model, and the validation dataset is used as a validation dataset during the training phase. We generate a fully new test dataset for synthetic datasets and evaluate $C$-$acc$ and $Dr$-$acc$.
We train all models with a learning rate $\alpha=0.00001$, a maximum batch size of 16 instances (less if GPU memory cannot fit 16 instances), and a maximal number of epochs equal to 1000 (we use early stopping and stop before 1000 epochs if the model starts overfitting the test dataset). 
For dCAM, we use $k=100$ (number of random permutations), a value that we empirically verified (due to lack of space, a detailed analysis of the effect of $k$ is in the full version of the paper).

\subsection{Classification Accuracy evaluation}
\label{sc:classifacc}

\begin{table*}[tb]
	\centering
	\scalebox{0.71}{
\begin{tabular}{|c|c|c|c||c|c|c|c|c|c|c||c|c|c||c|c|c|}
\hline
\rowcolor{Gray}
\multicolumn{4}{|c||}{{\bf \emph{Metadata}}} & \multicolumn{13}{|c|}{{\bf \emph{ $C$-$acc$ (averaged on 10 runs)}}} \\
\hline
\rowcolor{Gray}
\multicolumn{4}{|c||}{{\bf \emph{ }}} & \multicolumn{7}{|c||}{{\bf \emph{Baselines}}} & \multicolumn{3}{|c||}{{\bf \emph{c-Baselines}}} & \multicolumn{3}{|c|}{{\bf \emph{d-Baselines}}} \\
\hline
Datasets name   & $|\mathcal{C}|$ & $|T|$ & $D$ &\commentRed{RNN}&\commentRed{GRU}& \commentRed{LSTM} &MTEX& CNN  & ResNet & InceptionT. & cCNN & cResNet & cInceptionT. & dCNN & dResNet & dInceptionT. \\
\hline
AtrialFibrillation 		& 3	 & 640	& 2  		&\commentRed{0.66}	&\commentRed{0.70}	&\commentRed{0.66}	&{\bf0.72}	&0.41	&0.40	&0.64	&0.56	&0.53	&0.68	&0.49	&0.45	&0.61	\\
Libras   				& 15	 & 45		& 2  		&\commentRed{0.86}	&\commentRed{0.84}	&\commentRed{0.75}	&0.93	&{\bf0.96}	&{\bf0.96}	&0.82	&0.80	&0.82	&0.65	&0.91	&0.94	&0.66	\\
BasicMotions  			& 4	 & 100	& 2  		&\commentRed{0.68}	&\commentRed{0.87}	&\commentRed{0.83}	&0.91	&{\bf1.00}	&{\bf1.00}	&{\bf1.00}	&{\bf1.00}	&{\bf1.00}	&{\bf1.00}	&{\bf1.00}	&{\bf1.00}	&{\bf1.00}	\\
RacketSports  			& 4	 & 30		& 6  		&\commentRed{0.70}	&\commentRed{0.78}	&\commentRed{0.75}	&0.81	&0.94	&{\bf0.99}	&0.90	&0.95	&0.98	&0.85	&0.94	&0.98	&0.92	\\
Epilepsy   			& 4	 & 206	& 3  		&\commentRed{0.63}	&\commentRed{0.83}	&\commentRed{0.83}	&0.97	&{\bf1.00}	&{\bf1.00}	&0.97	&{\bf1.00}	&{\bf1.00}	&{\bf1.00}	&{\bf1.00}	&{\bf1.00}	&0.99	\\
StandWalkJump 		& 3	 & 2500 	& 4  		&\commentRed{0.60}	&\commentRed{0.80}	&\commentRed{0.90}	&0.66	&0.70	&0.66	&0.65	&0.83	&{\bf1.00}	&0.81	&0.95	&{\bf1.00}	&0.75	\\
UWaveGest.Lib. 		& 8	 & 315 	& 3  		&\commentRed{0.88}	&\commentRed{{\bf0.93}}	&\commentRed{0.88}	&{\bf0.93}	&0.88	&0.89	&0.89	&0.76	&0.74	&0.64	&0.84	&0.89	&0.83	\\
Handwriting  			& 26	 & 152	& 3  		&\commentRed{0.45}	&\commentRed{0.43}	&\commentRed{0.40}	&0.34	&0.83	&{\bf0.90}	&0.55	&0.42	&0.70	&0.38	&0.76	&0.89	&0.52	\\
NATOPS   			& 6  	 & 51		& 24		&\commentRed{0.87}	&\commentRed{0.91}	&\commentRed{0.82}	&0.91	&0.99	&{\bf1.00}	&0.95	&0.86	&0.89	&0.83	&0.97	&0.99	&0.91	\\
PenDigits   			& 10 & 8		& 2		&\commentRed{{\bf0.99}}	&\commentRed{{\bf0.99}}	&\commentRed{{\bf0.99}}	&{\bf0.99}	&{\bf0.99}	&{\bf0.99}	&{\bf0.99}	&{\bf0.99}	&{\bf0.99}	&{\bf0.99}	&{\bf0.99}	&{\bf0.99}	&{\bf0.99}	\\
FingerMovements  		& 2  	 & 50		& 28		&\commentRed{0.56}	&\commentRed{0.58}	&\commentRed{0.58}	&0.62	&0.70	&0.68	&0.71	&0.57	&0.63	&0.55	&{\bf0.72}	&0.71	&0.66	\\
Artic.WordRec.			& 25 & 144	& 9		&\commentRed{0.98}	&\commentRed{0.97}	&\commentRed{0.90}	&0.98	&{\bf0.99}	&{\bf0.99}	&0.93	&0.82	&0.94	&0.74	&0.98	&{\bf0.99}	&0.88	\\
HandMov.Dir. 			& 4  	& 400	& 10		&\commentRed{0.46}	&\commentRed{0.40}	&\commentRed{0.37}	&0.44	&0.44	&0.42	&{\bf0.51}	&0.34	&0.35	&0.40	&0.45	&0.44	&0.33	\\
Cricket   				& 12 & 1197	& 6		&\commentRed{0.98}	&\commentRed{0.90}	&\commentRed{0.80}	&0.95	&{\bf1.00}	&{\bf1.00}	&0.98	&0.94	&0.97	&0.87	&{\bf1.00}	&{\bf1.00}	&0.98	\\
LSST   				& 14 & 36		& 6		&\commentRed{0.54}	&\commentRed{0.53}	&\commentRed{0.52}	&0.53	&0.62	&{\bf0.66}	&0.40	&0.56	&0.59	&0.49	&0.62	&{\bf0.66}	&0.51	\\
Eth.Concentration 		& 4  	& 1751	& 3		&\commentRed{0.33}	&\commentRed{0.37}	&\commentRed{0.32}	&{\bf0.58}	&0.35	&0.36	&0.34	&0.36	&0.36	&0.36	&0.35	&0.39	&0.34	\\
SelfReg.SCP1 			& 2  	& 896	& 6		&\commentRed{{\bf0.91}}	&\commentRed{0.88}	&\commentRed{0.89}	&0.88	&0.86	&0.83	&0.87	&0.88	&0.84	&0.88	&0.86	&0.86	&0.88	\\
SelfReg.SCP2 			& 2  	& 1152	& 7		&\commentRed{0.58}	&\commentRed{0.57}	&\commentRed{0.60}	&0.58	&0.59	&0.58	&0.62	&0.60	&0.59	&0.59	&0.57	&0.60	&{\bf0.63}	\\
Heartbeat   			& 2  	& 405	& 61		&\commentRed{0.73}	&\commentRed{0.73}	&\commentRed{0.73}	&0.75	&0.83	&{\bf0.86}	&0.83	&0.76	&0.76	&0.76	&0.84	&{\bf0.86}	&0.83	\\
PhonemeSpectra  		& 39 & 217	& 39		&\commentRed{0.11}	&\commentRed{0.11}	&\commentRed{0.11}	&0.15	&0.31	&0.37	&0.27	&0.31	&0.33	&0.28	&0.33	&{\bf0.40}	&0.32	\\
EigenWorms  			& 5  	& 17984	& 6		&\commentRed{0.57}	&\commentRed{0.50}	&\commentRed{0.66}	&0.57	&0.90	&{\bf0.92}	&0.82	&0.71	&{\bf0.92}	&0.73	&{\bf0.92}	&{\bf0.92}	&0.81	\\
MotorImagery  		& 2  	& 3000	& 64		&\commentRed{0.53}	&\commentRed{0.59}	&\commentRed{0.58}	&0.59	&0.58	&0.57	&0.56	&0.56	&0.57	&0.56	&0.65	&{\bf0.68}	&0.66	\\
FaceDetection  		& 2  	& 62		& 144	&\commentRed{0.64}	&\commentRed{0.58}	&\commentRed{0.60}	&{\bf0.72}	&0.57	&0.59	&0.71	&0.55	&0.70	&0.70	&0.57	&0.61	&0.63	\\
\hline
\hline
\rowcolor{lGray}
\multicolumn{4}{|c||}{{\bf \emph{Mean}}} 			&0.662&0.686&0.672&0.717& 0.758& 0.766& 0.735& 0.701& 0.747& 0.684&0.770&\textbf{0.793}&0.723\\
\hline
\rowcolor{lGray}
\multicolumn{4}{|c||}{{\bf \emph{Rank}}} 			&8.26&7.65&8.73&6.39&4.73&4.13&6.08&7.30&5.73&7.73&4.56&\textbf{2.65}&6.56\\
\hline
\end{tabular}
}
\caption{\commentRed{$C$-$acc$ averaged accuracy for 10 runs 
over UCR/UEA datasets.}}
\vspace*{-0.7cm}
\label{result_Multi_Normal}
\end{table*}

We first evaluate the classification performance of our proposed approaches (denoted as $c$-Baselines and $d$-Baselines in Table~\ref{result_Multi_Normal}) and the different baselines (denoted as Baselines in Table~\ref{result_Multi_Normal}) over the UCR/UEA multivariate data series. 
We run each method ten times and report the average $C$-$acc$. 

\commentRed{We first observe that the recurrent models (RNN, GRU, LSTM) are less accurate by approximately 0.10 than CNN-based models (CNN, ResNet and InceptionTime). These results confirm the observations of previous works~\cite{10.1007/978-3-030-00937-3_25,DBLP:journals/corr/abs-1809-04356,DBLP:journals/corr/WangYO16,DBLP:journals/datamine/RuizFLMB21}}. 
We then observe that ResNet-based architecture performs better than CNN-based and InceptionTime-based architectures. 
Moreover, we note that, overall, dCNN and dResNet have a better $C$-$acc$ than CNN and ResNet, respectively. 
This observation confirms that our proposed architectures (dResNet, dCNN) do not result in any loss in accuracy; on the contrary, they are slightly more accurate than usual architectures (ResNet, CNN). 
We notice that dResNet is, on average, one rank higher than ResNet. 
Similar observations can be made when comparing dCNN and CNN.

\begin{figure}
	\centering
	\includegraphics[scale=0.49]{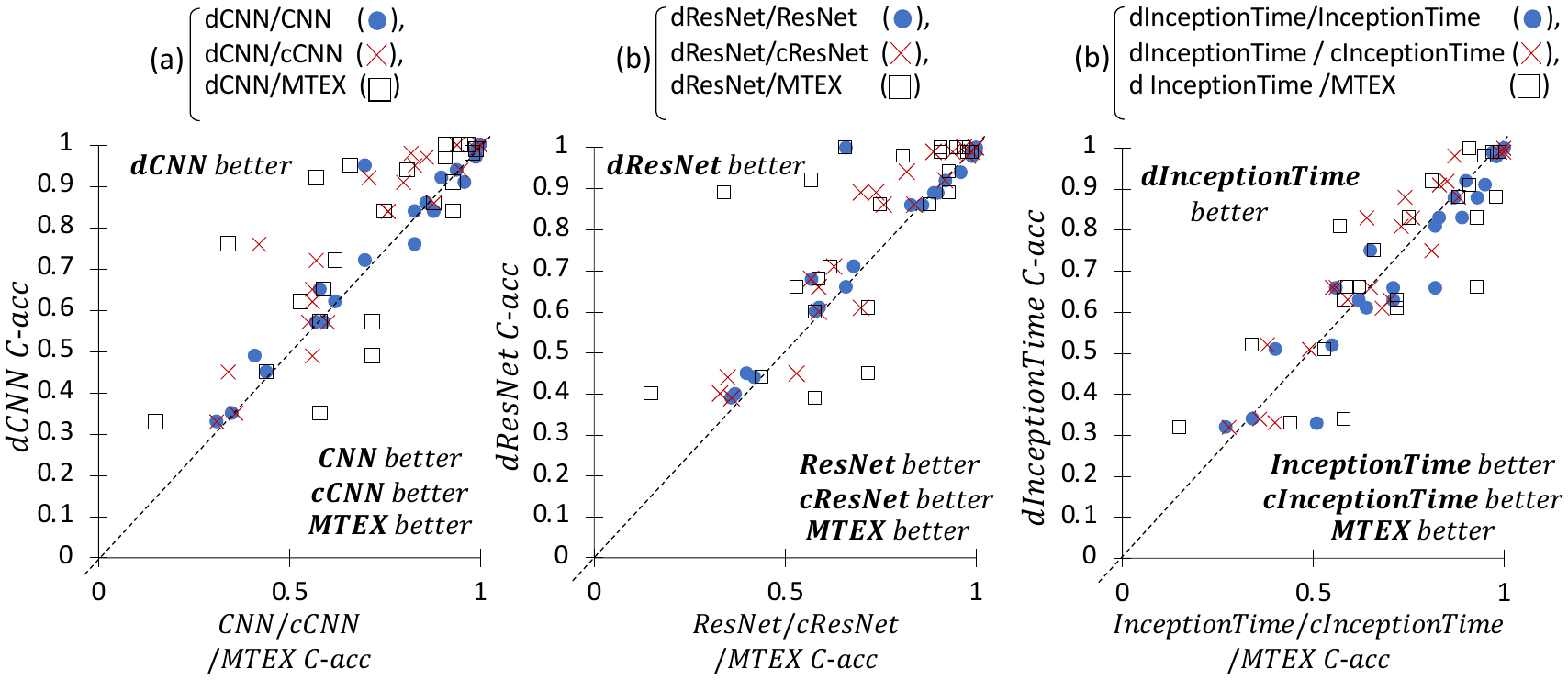}
	\vspace*{-0.7cm}
	\caption{$C$-$acc$ comparison of (a) dCNN with cCNN, CNN and MTEX, (b) dResNet with cResNet, ResNet and and MTEX, and (c) dInceptionTime with cInceptionTime, InceptionTime and MTEX on UCR/UEA datasets.}
	\label{UCRexpfig}
	\vspace*{-0.2cm}
\end{figure}

Moreover, Table~\ref{result_Multi_Normal} confirms that using cCNN baselines (or cResNet and cInceptionTime) implies a drop in classification accuracy. 
For instance, CNN architecture is 0.05 more accurate than cCNN architecture. 
Thus, $c$-Baselines cannot guarantee at least equivalent accuracy. 
Figure~\ref{UCRexpfig}(a) depicts the comparison between dCNN $C$-$acc$ (on the y-axis) and CNN/cCNN $C$-$acc$ (on the x-axis; CNN: blue circles; cCNN: red crosses). 
The dotted line corresponds to cases when both classifiers have the same accuracy. 
We observe that almost all cCNN $C$-$acc$ (red crosses) are above the dotted line, which shows that dCNN is more accurate for most datasets. 
Similarly, we observe that most of the CNN $C$-$acc$ (blue circles) are above the dotted lines, which means that dCNN is more accurate than CNN. 
The same observation can be made when examining Figure~\ref{UCRexpfig}(b), in which dResNet is compared with ResNet and cResNet.

However, the same observation is not true when comparing dInceptionTime with InceptionTime and cInceptionTime. 
Even though in Figure~\ref{UCRexpfig}(c) most of the red crosses are above the dotted line, indicating that dInceptionTime is most of the time more accurate than cInceptionTime, the blue circles are equally distributed above and under the dotted line. 
Thus, dInceptionTime is not more accurate than InceptionTime. 
The results in Table~\ref{result_Multi_Normal} also show that the averaged $C$-$acc$ across all datasets (as well as the averaged rank) is lower for dInceptionTime than for InceptionTime. 
Nevertheless, 
the performance of dInceptionTime is very close to that of InceptionTime. Thus, transforming the original architecture into one that supports dCAM does not penalize classification performance.

Finally, we observe that the accuracy of MTEX-CNN is lower than that of \textit{Baselines} and the \textit{d-Baselines}. 
We note that MTEX-CNN and cCNN have very similar performance (average accuracy of 0.71 and 0.70, and average rankings of 6.39 and 7.30). 
As we explained earlier (see Section~\ref{backprellimitation}), the MTEX-CNN architecture is divided into two blocks. 
The experiments demonstrate that the 2nd block cannot capture all discriminant features, and thus, cannot reach the accuracy of a traditional CNN. 
We conclude that MTEX-CNN is not as accurate as traditional architectures (such as CNN) or our proposed architectures (such as dCNN).

\subsection{Discriminant Features Identification}
\label{sc:explanacc}

\begin{table*}
\centering
	\scalebox{0.67}{
\begin{tabular}{|c|c|c|||c|c|c|c|c|c|||c|c||c||c|c|c||c|}
\hline
\rowcolor{Gray}
\multicolumn{3}{|c|||}{{\bf \emph{Datasets}}}    & \multicolumn{6}{|c|||}{{\bf \emph{$C$-$acc$ (averaged on 10 runs)}}} & \multicolumn{7}{|c|}{{\bf \emph{$Dr$-$acc$ (averaged on 50 instances)}}}  \\
\hline
\rowcolor{Gray}
       & &    &   & & & & &  & {\bf MTEX-grad} & {\bf CAM} & {\bf cCAM} & \multicolumn{3}{|c||}{{\bf dCAM}} & \\
\hline
Dataset name      & Type   & Dimensions & MTEX & ResNet & cResNet & dCNN  & dResNet & dInception &MTEX &ResNet& cResNet & dCNN  & dResNet & dInception & Random \\
\hline
\multirow{10}{*}{StarLightCurve} & \multirow{5}{*}{Type 1} & 10 &0.99	& 0.95  	&{\bf1.00} & {\bf1.00}& {\bf1.00}& {\bf1.00}&0.40 & 0.07*  & {\bf0.92}& 0.46   & 0.38    & 0.21   & 0.02 \\
       		&    							& 20 &0.99	& 0.71  	&{\bf1.00} & {\bf1.00}& {\bf1.00}& 0.98 	&0.38 & 0.02*  & {\bf0.92}& 0.38   & 0.45    & 0.36   & 0.01 \\
 		    	&    							& 40 &0.98	& 0.60  	&{\bf1.00} & 0.99 	& {\bf1.00}& 0.93 	&0.24 & 0.008* & {\bf0.94}& 0.28   & 0.42    & 0.39   & 0.005 \\
       		&    							& 60 &0.61	& 0.57 	&{\bf1.00} & 0.98 	& 0.99 	& 0.91 	&0.05 & 0.004* & {\bf0.92}& 0.23   & 0.24    & 0.13   & 0.003 \\
       		&    							&100&0.55	& 0.64 	&{\bf1.00} & 0.96 	& 0.97 	& 0.79 	&0.01 & 0.003* & {\bf0.92}& 0.2    & 0.26    & 0.10   & 0.002 \\
\cline{2-16}
       & \multirow{5}{*}{Type 2} 					& 10 &0.58	& 0.71  	& 0.53 	& {\bf1.00}& {\bf1.00}& 0.93 	&0.15 & 0.0256*& 0.025  & 0.26   & {\bf0.43}& 0.10   & 0.021 \\
       		&    							& 20 &0.55	& 0.61  	& 0.55 	& 0.98 	& {\bf1.00}& 0.70 	&0.04 & 0.016* & 0.01   & 0.28   & {\bf0.43}& 0.05   & 0.01 \\
       		&    							& 40 	&0.56	& 0.58  	& 0.51 	& {\bf0.88}& 0.58 	& 0.56 	&0.07 & 0.0068*& 0.006  & {\bf0.20}& 0.05   & 0.03   & 0.005 \\
       		&    							& 60 	&0.53	& 0.55  	& 0.53 	& {\bf0.64}& 0.59 	& 0.55 	&0.008& 0.0058*& 0.005  & {\bf0.01}& 0.003  & 0.009  & 0.003 \\
		    	&    							&100&0.52	& 0.59  	& 0.5 	& 0.59 	& 0.56 	& {\bf0.60}&0.01 & 0.0024*& 0.002  & 0.003  & 0.004  &{\bf0.02}& 0.002 \\
\hline
\hline
\multirow{10}{*}{ShapesAll} & \multirow{5}{*}{Type 1} 	& 10 &{\bf1.00} & {\bf1.00}& {\bf1.00}& {\bf1.00}& {\bf1.00}& {\bf1.00}&0.60 & 0.09*  & {\bf0.79}& 0.66   & 0.7    & 0.55    & 0.02 \\
		     &     							& 20 &{\bf1.00} & 0.86  	& {\bf1.00}& {\bf1.00}& {\bf1.00}& 0.99 	&0.31 & 0.03*  & {\bf0.74}& 0.56   & 0.66   & 0.51    & 0.011 \\
		     &     							& 40 &0.85	& 0.65  	& {\bf1.00}& {\bf1.00}& {\bf1.00}& {\bf1.00}&0.20 & 0.008* & {\bf0.88}& 0.45   & 0.74   & 0.76    & 0.005 \\
		     &     							& 60 &0.83	& 0.65  	& {\bf1.00}& {\bf1.00}& {\bf1.00}& 0.96 	&0.50 & 0.005* & 0.65   & 0.44   & 0.72    & {\bf0.79}& 0.003 \\
		     &     							&100&0.70	& 0.57  	& {\bf1.00}& 0.98 	& {\bf1.00}& 0.85 	&0.002& 0.003* & {\bf0.83}& 0.31   & 0.49   & 0.48    & 0.002 \\
\cline{2-16}
		     & \multirow{5}{*}{Type 2} 				& 10 &0.60	& 0.82 	& 0.54 	&{\bf1.00} & {\bf1.00}& 0.93 	&0.02 & 0.0467*& 0.04   & {\bf0.63}& 0.50   & 0.32   & 0.02 \\
		     &     							& 20 &0.54	& 0.57  	& 0.52 	&{\bf1.00} & {\bf1.00}& 0.89 	&0.04 & 0.0132*& 0.013  & 0.50   & {\bf0.73}& 0.40   & 0.01 \\
		     &     							& 40 &0.59	& 0.60  	& 0.52 	&{\bf0.90} & 0.72 	& 0.73 	&0.02 & 0.005* & 0.005  & {\bf0.40}& 0.20   & 0.36   & 0.005 \\
		     &     							& 60 &0.57	& 0.59  	& 0.51 	& 0.65 	& 0.61 	& {\bf0.72}&0.06 & 0.0037*& 0.003  & 0.22   & 0.34   & {\bf0.46}& 0.003 \\
		     &     							&100&0.52	& {\bf0.59}& 0.50 	& 0.55 	& 0.58 	& 0.55 	&0.04 & 0.0027*& 0.002  & 0.005  & 0.02   & {\bf0.05}& 0.002 \\
\hline
\hline
\rowcolor{lGray}
\multicolumn{3}{|c|}{{\bf \emph{Rank}}}         		&3.95& 3.9  & 3  & 1.65 & {\bf 1.6} & 2.85 &3.85& 4.45 & 3  & 2.6  & {\bf 2.15}  & 2.75  & \\
\hline
\end{tabular}
}
\caption{$C$-$acc$ and $Dr$-$acc$ averaged accuracy for 10 runs 
over synthetic datasets.}
\label{result_Multi_Normal_synth}
\vspace*{-0.7cm}
\end{table*}

\begin{figure}
	\centering
	\includegraphics[scale=0.67]{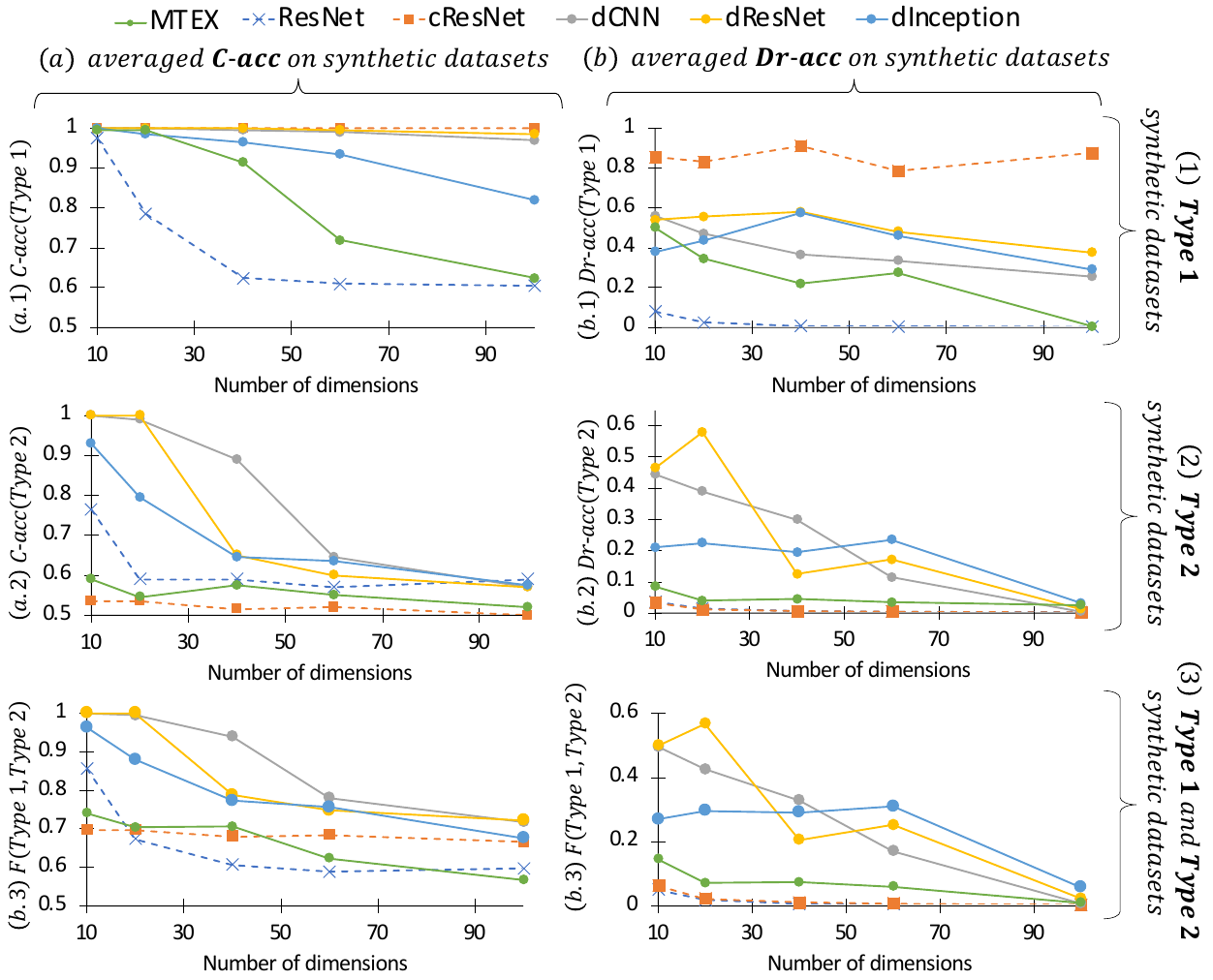}
	\vspace*{-0.75cm}
	\caption{Evaluation of influence of number of dimensions on our approaches and the baselines $C$-$acc$ and $Dr$-$acc$. }
	\label{synthexpfig}
	\vspace*{-0.5cm}
\end{figure}

We now evaluate the classification accuracy ($C$-$acc$) and the discriminant features identification accuracy ($Dr$-$acc$) on synthetically built datasets. 
Table~\ref{result_Multi_Normal_synth} depicts both $C$-$acc$ and $Dr$-$acc$ on $Type$ $1$ and $2$ datasets, when varying the number of dimensions from 10 to 100. 
In this experiment, we keep as baselines only ResNet and cResNet, which are the most accurate methods among all other baselines.

Overall, 
all methods have better performance (both $C$-$acc$ and $Dr$-$acc$) on $Type$ $1$ datasets than on $Type$ $2$. 
This was expected: discriminant features located in single dimensions are easier to find than discriminant features that depend on several dimensions. 

We then notice that for low dimensional ($D$=10) datasets, ResNet, dResNet, dCNN, and dInceptionTime are performing nearly perfect $\emph{C-acc}$. 
Moreover, ResNet and MTEX-CNN are performing well for low-dimensional data series but start to fail for a more significant number of dimensions. 
While the drop is already significant for the $Type$ $1$ dataset built from the StarLightCurve dataset, it is even stronger for $Type$ $2$ datasets, for which ResNet fails to classify instances with a number of dimensions $D \geq 20$. 
On the contrary, dCNN, dResNet, and dInceptionTime, which use the random permutations in the input, are not sensitive to the number of dimensions and have an almost perfect $C$-$acc$ for most of $Type$ $1$ datasets. 
We observe a $C$-$acc$ drop for dCNN, dResNet and dInceptionTime as dimensions increase for $Type$ $2$ datasets.
However, this drop is significantly less pronounced than that of ResNet.
Overall, dCNN, dResNet, and dInceptionTime, which have on average the three highest ranks, are the most accurate methods.

Regarding cResNet, although it achieves a nearly perfect $\emph{C-acc}$ for $Type$ $1$ datasets, we observe that it fails to classify correctly instances of $Type$ $2$ datasets. 
As explained in Section~\ref{sec:prelim}, the input data structure is not rich enough to allow comparisons among dimensions, which is the main way to find discriminant features between the two classes of $Type$ $2$ datasets. 
We also observe that MTEX-CNN fails to classify instances of $Type$ $2$ datasets. 
Thus, this architecture does not correctly detect the discriminant features across different dimensions.
Overall, Figure~\ref{synthexpfig}(a) shows that dCNN, dResNet and dInceptionTime are equivalent to cResNet for $Type$ $1$ (Figure~\ref{synthexpfig}(a.1)), outperforming all the baselines for $Type$ $2$ (Figure~\ref{synthexpfig}(a.2)), and in general are better than the baselines (ResNet and cResNet) for both types (Figure~\ref{synthexpfig}(a.3) with $F(Type\text{ }1,Type\text{ }2) = \frac{2*C\text{-}acc(Type\text{ }1)*C\text{-}acc(Type\text{ }2)}{C\text{-}acc(Type\text{ }1)+C\text{-}acc(Type\text{ }2)}$).

We now compare the different methods using the $Dr$-$acc$ measure. 
We observe that the baseline cCAM (computed with cCNN) is outperforming CAM (computed with ResNet) and dCAM (with all of dCNN, dResNet and dInceptionTime) for $Type$ $1$ datasets. 
This is explained by the fact that these classes can be discriminated by treating dimensions independently. 
Thus, cCAM (with no comparisons between dimensions) is naturally the best solution. 
Nevertheless, as $Type$ $2$ datasets require comparisons among dimensions to discriminate the classes, cCAM fails on them, with a $Dr$-$acc$ very similar to the one of a random classifier. 
This confirms that such a baseline cannot be considered as a general solution for multivariate data series classification.
We also observe that $Dr$-$acc$ of the explanation method of MTEX-CNN (MTEX-grad) is lower than dCAM for $Type$ $1$ and close to $Dr$-$acc$ of cCAM for $Type$ $2$, meaning that it cannot identify discriminant features of $Type$ $2$ datasets.

We then compare CAM and dCAM (used with dCNN/dResNet/ dInceptionTime). 
Figure~\ref{synthexpfig}(b) shows that dCAM significantly outperforms CAM, and that $Dr$-$acc$ reduces for all models as the number of dimensions increases. 
Nevertheless, $Dr$-$acc$ of dCAM remains relatively high for both $Type$ $1$ (Figure~\ref{synthexpfig}(b.1)) and $Type$ $2$ (Figure~\ref{synthexpfig}(b.2)) datasets (for less than 60 dimensions).

This result demonstrates the superiority of dCAM over state-of-the-art methods. 
Besides, the average ranks in Table~\ref{result_Multi_Normal_synth}
indicate 
that dCAM computed from ResNet has the highest rank of $2.15$.

\commentRed{

\subsection{Influence of $k$}

This section analyzes the influence of the number of permutations $k$ on the discriminative features identification accuracy ($Dr$-$acc$). 
We compute the $Dr$-$acc$ for 20 different instances for which dCAM is computed using a value of $k$ between $1$ and $400$. 
We randomly select the 20 instances from the 9 ShapesAll datasets, $Type$ $1$ and $Type$ $2$. 
(We excluded the $Type$ $2$ ShapesAll dataset with 100 dimensions, because no model trained on this dataset leads to reasonably accurate results: see Table~\ref{result_Multi_Normal_synth}.)
Figure~\ref{convergence_k}(a.1) for $Type$ $1$ datasets and (a.2) for $Type$ $2$ datasets depicts the evolution of $Dr$-$acc$ (on average for the 20 instances) as $k$ increases, for dCNN, dResNet and dInception Time. 
The results show that the model architecture influences convergence speed, and that convergence speed reduces as the number of dimensions increases. 
Figure~\ref{convergence_k}(b) shows that the number of permutations needed to reach 90 percents of the best $Dr$-$acc$ is greater when $D$ is higher. 
The latter holds for $Type$ $1$ (Figure~\ref{convergence_k}(b.1)) and $Type$ $2$ datasets (Figure~\ref{convergence_k}(b.2)). 
Overall, we notice that the dCAM computation with dResNet and dInceptionTime converges faster than dCNN. 
Studying deep neural network architectures 
that could 
reduce the number of permutations needed to reach the maximum $Dr$-$acc$ is an open research problem. 

\begin{figure}
	\centering
	\includegraphics[scale=0.50]{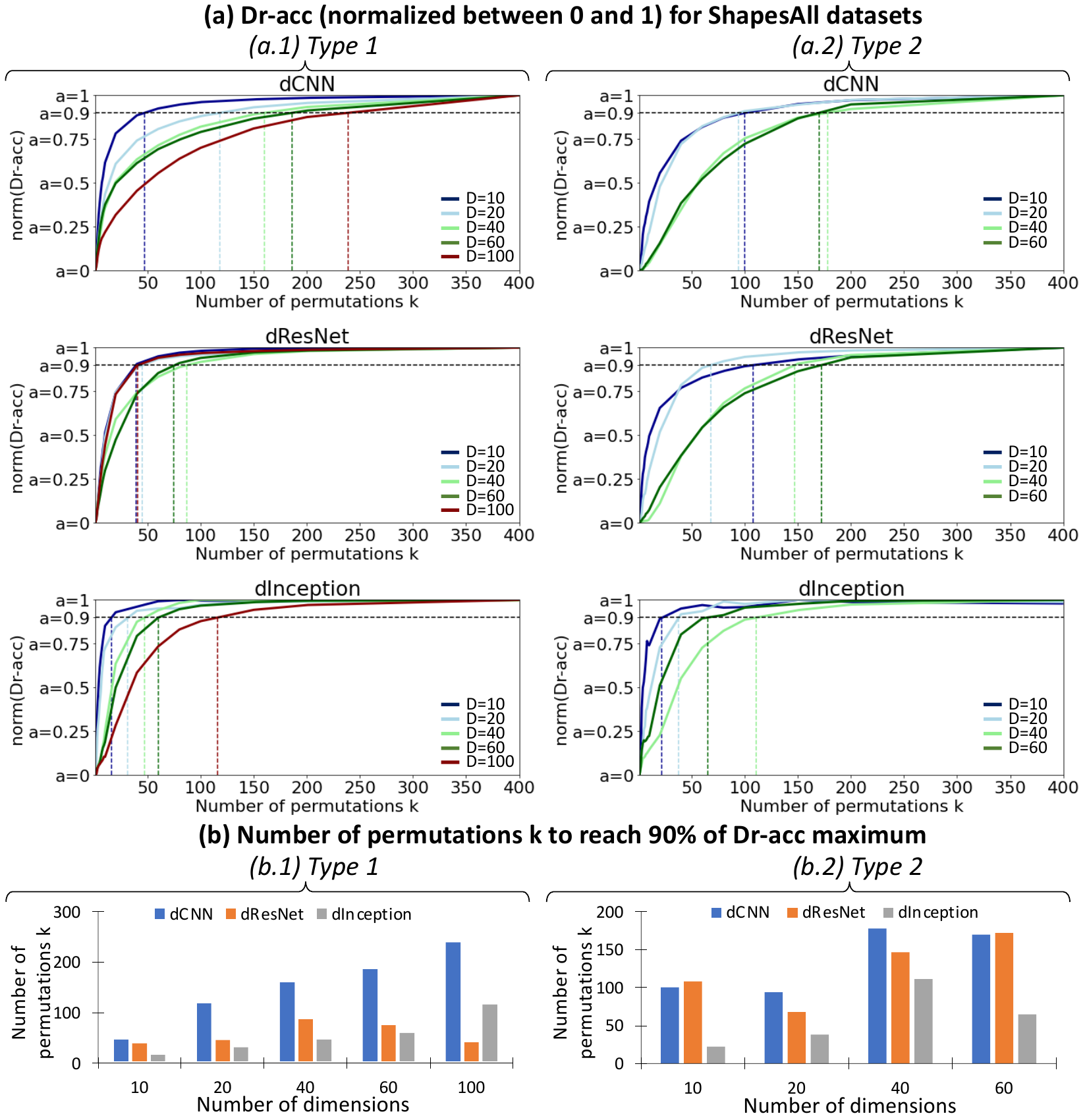}
	\vspace*{-0.4cm}
	\caption{\commentRed{Influence of $k$ on $Dr$-$acc$ for ShapesAll datasets.}}
	\label{convergence_k}
	\vspace*{-0.5cm}
\end{figure}

}

\subsection{$C$-$acc$ versus $Dr$-$acc$}
\label{sc:parameteranalysis}

\begin{figure}
	\centering
	\includegraphics[scale=0.67]{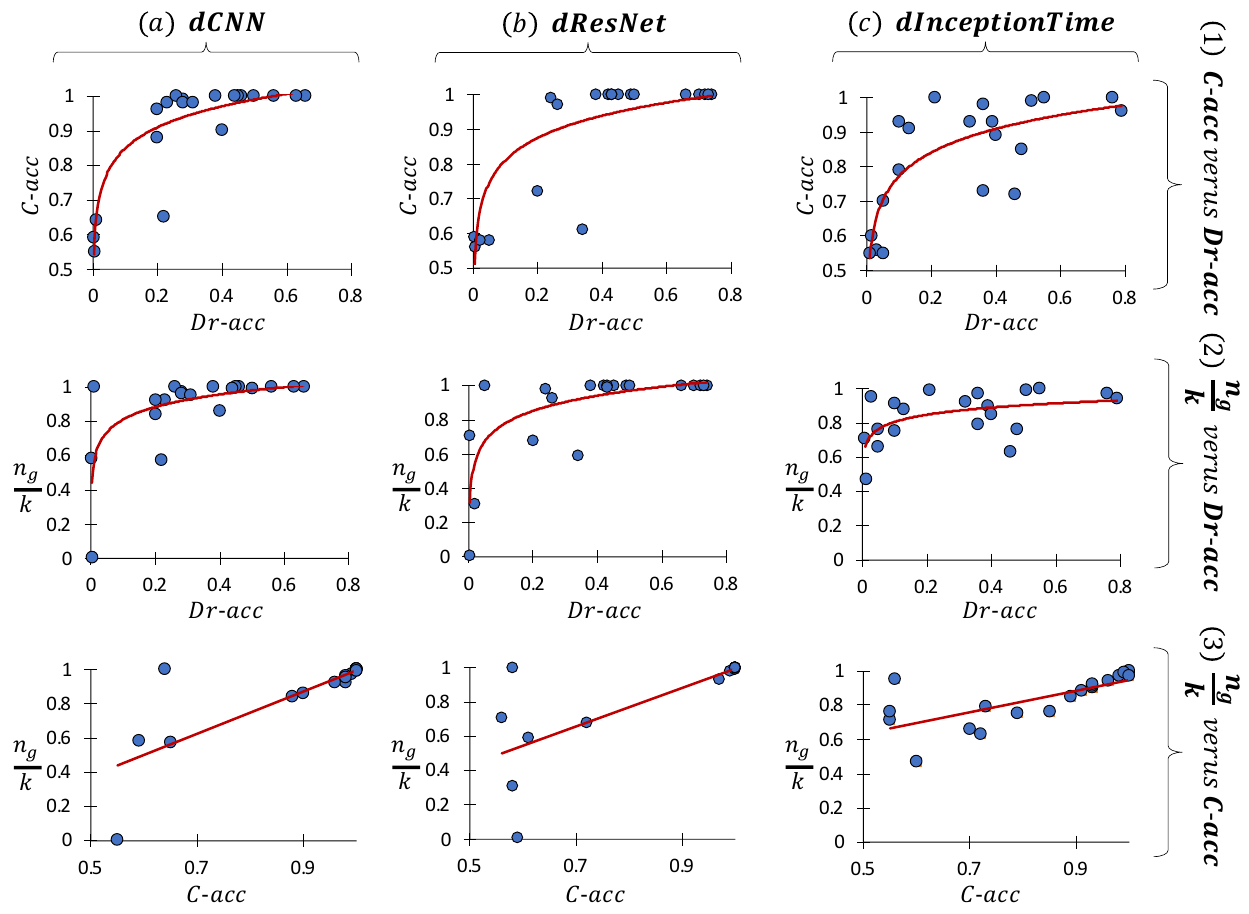}
	\vspace*{-0.75cm}
	\caption{{Evaluation of $C$-$acc$, $Dr$-$acc$, and ratio between number of permutations $k$ and number of permutations correctly classified $n_g$, for dCNN, dResNet and dInceptionTime.}}
	\label{synthexpfig_param}
	\vspace*{-0.3cm}
\end{figure}

In this section, we 
first analyze the relation between $C$-$acc$ and $Dr$-$acc$. 
We then evaluate the impact that $C$-$acc$ has on the number of permutations that have been correctly classified $n_g$.
We finally evaluate the impact that $n_g$ has on $Dr$-$acc$. 

Figure~\ref{synthexpfig_param}(1) depicts the relation between $C$-$acc$ and $Dr$-$acc$ for dCNN (Figure~\ref{synthexpfig_param}(a.1)), dResNet (Figure~\ref{synthexpfig_param}(b.1)) and dInceptionTime (Figure~\ref{synthexpfig_param}(c.1)) for all synthetic datasets. 
Note that all methods have a logarithmic relation (dotted red line) between $Dr$-$acc$ (x-axis) and $C$-$acc$ (y-axis). 
This confirms that the accuracy of the trained model has a significant impact on discriminant feature identification. 

Figure~\ref{synthexpfig_param}(3) depicts on the y-axis the ratio of correctly classified permutations ($n_g$) among all permutations ($k$) versus the $C$-$acc$ (on the x-axis). 
In this case, for all of dCNN (Figure~\ref{synthexpfig_param}(a.3)), dResNet (Figure~\ref{synthexpfig_param}(b.3)) and dInceptionTime (Figure~\ref{synthexpfig_param}(c.3)), we observe that there exists a linear relationship for $C$-$acc$ between 0.7 and 1. 
This means that $n_g$ will be greater when the model is more accurate. 
Nevertheless, for $C$-$acc$ between 0.5 and 0.7, we observe a high variance for $n_g/k$. 
Thus, an inaccurate model may still lead to a high $n_g$. 
Finally, Figure~\ref{synthexpfig_param}(2) depicts the relation between $n_g/k$ (on the y-axis) and $Dr$-$acc$ (on the x-axis). 
We observe a similar relationship between $C$-$acc$ and $Dr$-$acc$, which means that a low $n_g$ may lead to inaccurate discriminant features identification.

\commentRed{As hypothesized in Section~\ref{sec:solution}, the experimental results confirm that an inaccurate model (for all of dCNN, dResNet, and dInceptionTime) cannot be used to identify discriminant features. 
Moreover, since in a real use-case it is not possible to measure $Dr$-$acc$, we can use $n_g/k$ to estimate the discriminant feature identification accuracy. 
Even though Figure~\ref{synthexpfig_param}(2) demonstrates that a high $n_g/k$ does not always lead to a high $Dr$-$acc$, in practice, we can safely assume that a low $n_g/k$ will most probably correspond to a low $Dr$-$acc$. Therefore, such measure can be used as a proxy for the estimation of the explanation quality}.

\subsection{Execution time evaluation}
\label{sc:executiontime}

\begin{figure}
	\centering
	\includegraphics[scale=0.75]{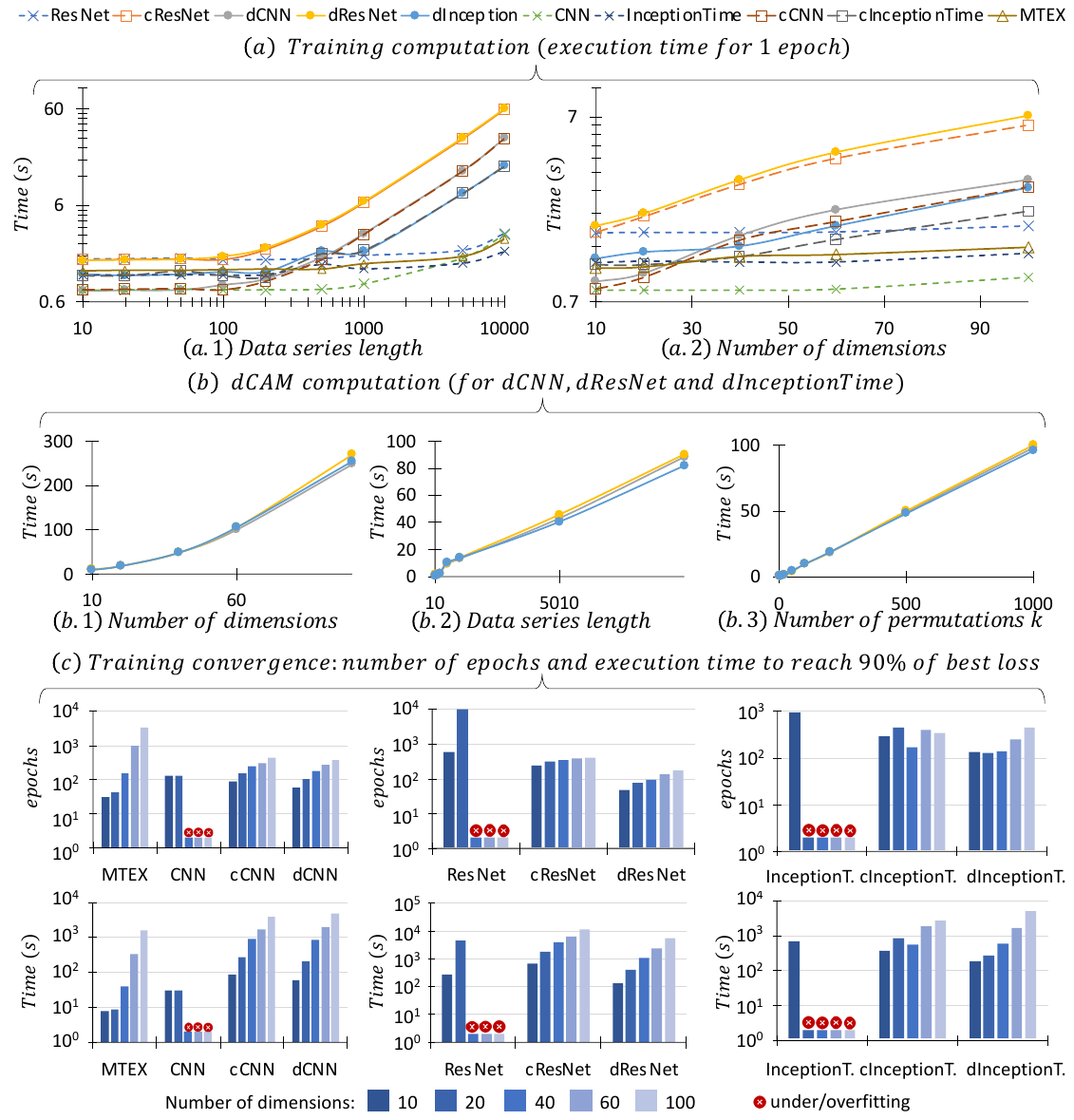}
	\vspace*{-0.65cm}
	\caption{\commentRed{Execution time (seconds) for (a) training computations when we vary (a.1) data series length and (a.2) number of dimensions. (b) Execution time for dCAM computation 
	when we vary (b.1) number of dimensions, (b.2) data series length, (b.3) number of permutations $k$, (c) training time to reach convergence (90\% of the best loss).}}
	\label{execTime}
	\vspace*{-0.3cm}
\end{figure}

\begin{figure*}
	\centering
	\includegraphics[scale=0.68]{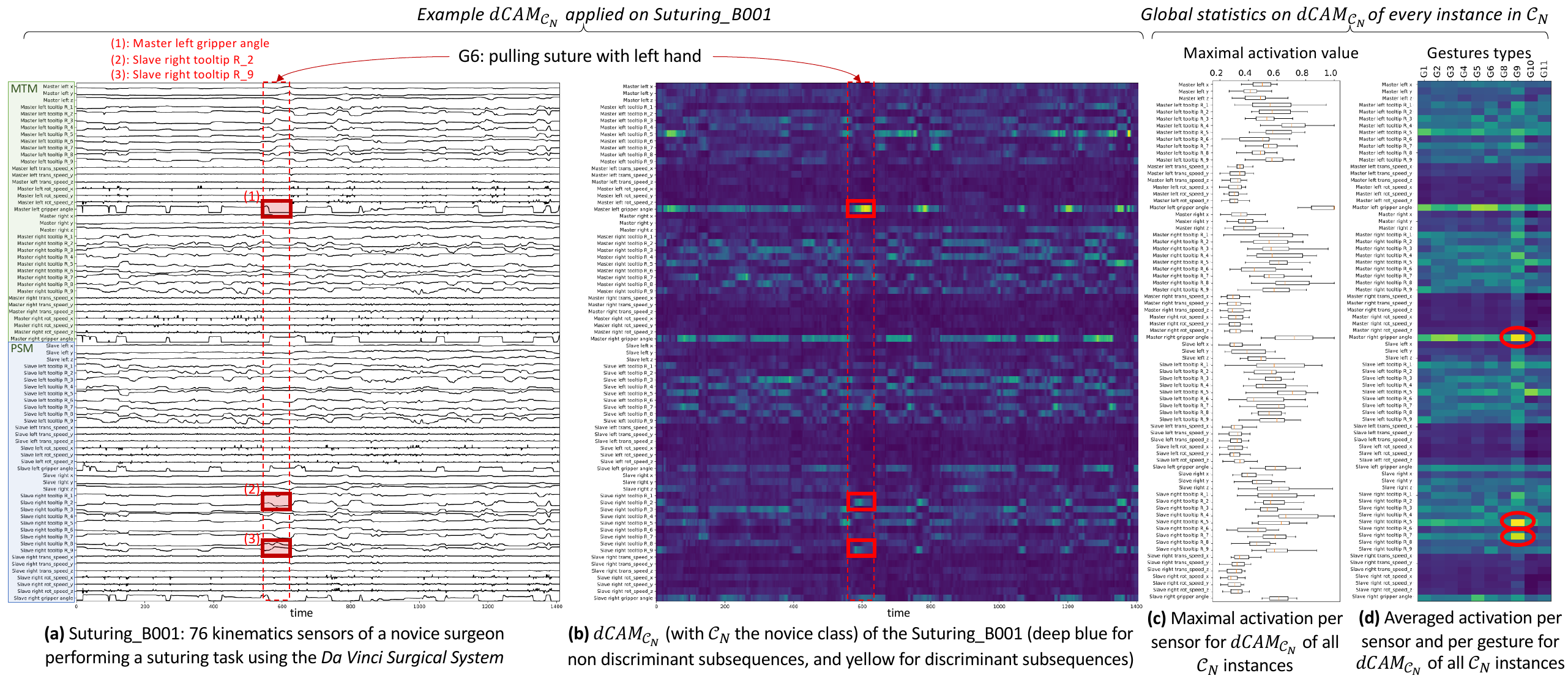}
	\vspace*{-0.3cm}
	\caption{Example of the result of dCAM on a multivariate data series of the JIGSAWS dataset. (a) the depicted data series corresponds to a novice surgeon performing a suture operation (joined with the corresponding dCAM in (b)). General statistical results over the entire novice class $\mathcal{C}_N$ are depicted such as (c) box-plots of the maximal activation value per sensor and (d) the averaged activation per sensor per gesture performed.}
	\label{surgery}
	\vspace*{-0.3cm}
\end{figure*}

In this section, we evaluate the execution time of our proposed approaches and the baselines. Figure~\ref{execTime}(a) depicts the training execution time (for one epoch) when we vary the data series length with a constant number of dimensions fixed to 10 (Figure~\ref{execTime}(a.1)), and when we vary the number of dimensions with a constant data series length fixed to 100 (Figure~\ref{execTime}(a.2)). 
\commentRed{In these two experiments, we use a batch size of 4 for all models.}
Overall, CNN and InceptionTime-based architectures are faster than ResNet-based architectures, and CNN, ResNet, InceptionTime, and MTEX-CNN are faster when the number of dimensions and the data series length is increasing. 
Nevertheless, both dCNN/dResNet/dInceptionTime and cCNN/cResNet/cInceptionTime require the same 
training time.

\commentRed{We now evaluate the execution time and the number of epochs required to train our proposed approaches and the baselines. Figure~\ref{execTime}(c) depicts the time (in seconds) and the number of epochs to reach 90\% of the best loss (on the test set) for $Type$ $1$ ShapesAll datasets varying the number of dimensions between 10 and 100. We use for all models a batch size of 16. The red dot indicates that a model is either overfitted or underfitted (i.e., the loss for the first epoch is approximately equal to the best loss). We observe that cCNN/cResNet/cInceptionTime and dCNN/dResNet/dInceptionTime require the same amount of time to be trained, but traditional baselines require more epochs than the proposed d-methods. Thus, training time for ResNet is longer than dResNet for $D=10$ and $D=20$. 
}

Finally, we measure the execution time required to compute dCAM (for dCNN, dResNet, and dInceptionTime), when we vary the number of dimensions with a constant data series length fixed to 400 (Figure~\ref{execTime}(b.1)), when we vary the data series length with a constant number of dimensions fixed to 10 (Figure~\ref{execTime}(b.2)), and when we vary the number of permutations (Figure~\ref{execTime}(b.3)). 
Note that the dCAM execution times are very similar for the three types of architectures. 
Moreover, the execution time increases super-linearly with the number of dimensions but is linear to the data series length and the number of permutations $k$.

\subsection{Use Case: Surgeon skills explanation}
\label{sec:surgery}

We now illustrate the applicability of our method to a real-world use case. 
In this use case, we train our dCNN network on the JIGSAWS dataset~\cite{Gao2014JHUISIGA} to identify novice surgeons, based on kinematic data series when performing \emph{surgical suturing} tasks (i.e., wound stitching) using robotic arms and surgical grippers. 

\noindent{\bf [Dataset]}
The data series are recorded from the $Da Vinci Surgical System$. The multivariate data series are composed of $76$ dimensions (an example of multivariate data series is depicted in Figure~\ref{surgery}(a)). Each dimension corresponds to a sensor (with an acquisition rate of $30$ Hz). 
The sensors are divided into four groups: patient-side manipulators (left and right PSMs: green rectangle in Figure~\ref{surgery}(a) top left), and left and right master tool manipulators (left and right MTMs: blue rectangle in Figure~\ref{surgery}(a) bottom left). 
Each group contains 19 sensors. 
These sensors are: 
3 variables for the Cartesian position of the manipulator, 
9 variables for the rotation matrix, 
6 variables for the linear and angular velocity of the manipulator, and 
1 variable for the gripper angle. 

To perform a suture, the surgeons perform different gestures (11 in total). 
For example, $G1$ refers to reaching for the needle with the right hand, while $G11$ refers to dropping the suture at the end and moving to end points. 
Each gesture corresponds to a specific time segment of the dataset, involving all sensors. 
For example, the dotted red rectangle in Figure~\ref{surgery}(a) represents gesture $G6$: pulling the suture with the left hand.
Surgeons that reported having more than 100 hours of experience are considered experts, 
surgeons with 10-100 hours are considered intermediate, and surgeons with less than 10 hours are labeled as novices.
We have 19 multivariate data series in the novice class, denoted as $\mathcal{C}_N$, 10 in the intermediate, $\mathcal{C}_i$, 10 multivariate data series in the expert class, $\mathcal{C}_E$.
More information on this dataset can be found in~\cite{Gao2014JHUISIGA}. 

\noindent{\bf [Training]}
For the training procedure, we use 80\% of the dataset (randomly selected from the three classes) for training.
The rest 20\% of the dataset is used for validation and early stopping. 
Since the instances do not have the same length, we use batches composed of one instance when training the models in the GPU. 

\noindent{\bf [Evaluation]}
Similar to what has been reported in previous work~\cite{10.1007/978-3-030-00937-3_25}, we achieve 100\% accuracy on the train and test datasets (for ten different randomly selected train and test sets). 
We proceed to compute the $dCAM_{\mathcal{C}_N}$ for every instance of the novice class $\mathcal{C}_N$. 
The $dCAM_{\mathcal{C}_N}$ of the multivariate data series named $Suturing$\_$B001$ (Figure~\ref{surgery}(a)) is displayed in Figure~\ref{surgery}(b). 
In the latter, the deep blue color indicates low activated subsequences (i.e., non-discriminant of belonging to the novice class $\mathcal{C}_N$), while the yellow color is pointing to highly activated subsequences. 
First, we note that some groups of sensors (dimensions) are more activated than others. 
In Figure~\ref{surgery}(b), the left and right "MTM gripper angles" are the most activated sensors. 
Figure~\ref{surgery}(c), which depicts the box-plot of the maximal activated values per sensor, confirms that in the general case, MTM gripper angles, as well as the MTM and PSM tooltip rotation matrices (three of these sensors are highlighted in red in Figure~\ref{surgery}(a)), are the most discriminant sensors. 
On the contrary, linear and angular speeds are not discriminant and hence cannot explain the novice class $\mathcal{C}_N$.

\commentRed{As explained in Section~\ref{sec:discussion}, we now extract global explanations at the scale of the dataset. 
We compute the dCAM for each instance, and we then extract global statistics on the sensors, i.e., aggregated over all instances}. Figure~\ref{surgery}(d) depicts the averaged activation per sensor per gesture.
Overall, $dCAM_{\mathcal{C}_N}$ identifies gesture $G9$ (using the right hand to help tighten the suture) as a discriminant gesture, because of the discriminant subsequences present in the sensors "right MTM gripper angle", "$5^{th}$ element", and "$7^{th}$ element" (marked with red ovals in Figure~\ref{surgery}(d)).
These three identified sensors (dimensions) are relevant to the right PSM tooltip rotation matrix and are important for the suturing process.
Similarly, we observe that gesture $G6$ (i.e., pulling suture with left hand) is discriminant, and activated the most by the "left NTM gripper angle" sensor. 
\commentRed{This result is consistent with a previous study~\cite{10.1007/978-3-030-00937-3_25}, which also identified gesture $G6$ as discriminant of belonging to the novice class. 
Nevertheless, this previous study was using CAM to only highlight the \emph{time interval} corresponding to gesture $G6$. 
On the contrary, dCAM provides more accurate (and useful) information: it does not only identify the discriminant gesture $G6$, 
but also the \emph{discriminant sensors}. This allows the analysts to recognize exactly what aspects of the particular gesture are problematic.}

\noindent{\bf [Summary]} 
The application of 
dCAM 
in the robot-assisted surgeon training use case demonstrated its effectiveness.
Our approach was able to provide meaningful explanations for the classification decisions, based on specific gestures (subsequences), and specific sensors (dimensions) that describe particular aspects of these gestures, i.e., the positioning and rotation angles of the tip of the stitch gripper.
Such explanations can help surgeons to improve their skills.

\section{Conclusions} 
\label{sec:concl}

Even though data series classification using deep learning has attracted a lot of attention, existing techniques for explaining the classification decisions fail for the case of multivariate data series. 
We described a novel approach, dCAM, based on CNNs, which detects discriminant subsequences within individual dimensions of a multivariate data series. 
The experimental evaluation with synthetic and real datasets demonstrates the 
superiority of our approach. 


\vspace*{0.2cm}
\noindent{\bf \large Acknowledgments}
Work supported by 
EDF R\&D and ANRT French program, and HPC resources
from GENCI-IDRIS (Grants 2020-101471 and 2021-101925), and NVIDIA Corporation for the Titan Xp GPU donation used in this research.

\bibliographystyle{ACM-Reference-Format}
\bibliography{anomalies}

\end{document}